%% file: main.tex
\documentclass[conference]{IEEEtran}
\input{macros.tex}
\newcommand{\Varun}[1]{\textcolor{cyan}{[\textbf{Varun}: #1]}}
\newcommand{\Brian}[1]{\textcolor{magenta}{[\textbf{Brian}: #1]}}

\AtBeginDocument{%
  \providecommand\BibTeX{{%
    \normalfont B\kern-0.5em{\scshape i\kern-0.25em b}\kern-0.8em\TeX}}}

\usepackage{accsupp}
\providecommand*{\napprox}{%
  \BeginAccSupp{method=hex,unicode,ActualText=2249}%
  \not\approx
  \EndAccSupp{}%
}

\begin{document}

%
\title{Rearchitecting Classification Frameworks For Increased Robustness}
\author{Varun Chandrasekaran\IEEEauthorrefmark{2}, Brian Tang\IEEEauthorrefmark{2}, Nicolas Papernot\IEEEauthorrefmark{1}\IEEEauthorrefmark{4},Kassem Fawaz\IEEEauthorrefmark{2},Somesh Jha\IEEEauthorrefmark{2}, Xi Wu\IEEEauthorrefmark{3}\vspace*{0.15cm} \\ 
University of Toronto\IEEEauthorrefmark{1}, Google \IEEEauthorrefmark{3},Vector Institute\IEEEauthorrefmark{4}, University of Wisconsin-Madison\IEEEauthorrefmark{2}}







\maketitle
\thispagestyle{plain}
\pagestyle{plain}

\input{Contents/0_abstract.tex}
\input{Contents/1_introduction.tex}
\input{Contents/2_background.tex}
\input{Contents/3_invariants.tex}
\input{Contents/5_general_framework.tex}
\input{Contents/4_motivation.tex}

\input{Contents/6_experiments.tex}
\input{Contents/7_discussion.tex}
\input{Contents/8_related_work.tex}
\input{Contents/9_conclusion.tex}
\bibliographystyle{IEEEtran}
\bibliography{main}

\appendix
\input{Contents/appendix.tex}

%



\end{document}

%% file: macros.tex
\usepackage{amsfonts}
\usepackage{amsmath}
\usepackage{multirow}
\usepackage{xspace}
\usepackage{subfig}
\usepackage{xcolor}
\usepackage{soul}
\usepackage{algorithm}
\usepackage[noend]{algpseudocode}
\usepackage{booktabs}
\usepackage{subfig}
\usepackage{mathtools}
\usepackage{float}
\usepackage{wasysym}
\usepackage{url}
\newcommand{\lidar}{LiDAR\xspace}
\newcommand{\N}{\ensuremath{\mathcal{N}}}

\DeclareMathOperator{\argmax}{{\rm argmax}}

\DeclareMathOperator{\R}{{\mathbb R}}

\newcommand{\sign}{\ensuremath{\text{sign}}}

\newtheorem{theorem}{Theorem}

\newcommand{\ie}{\textit{i.e.}\@\xspace}
\newcommand{\eg}{\textit{e.g.,}\@\xspace}
\newcommand{\etal}{\textit{et al.}\@\xspace}

%% file: Contents/0_abstract.tex
\begin{abstract}


While generalizing well over natural inputs, neural networks are vulnerable to adversarial inputs. Existing defenses against adversarial inputs have largely been detached from the real world. These defenses also come at a cost to accuracy. Fortunately, there are invariances of an object that are its salient features; when we break them it will {\em necessarily change} the perception of the object. We find that applying invariants to the classification task makes robustness and accuracy feasible together. Two questions follow: how to extract and model these invariances? and how to design a classification paradigm that leverages these invariances to improve the robustness accuracy trade-off? The remainder of the paper discusses solutions to the aformenetioned questions.

\end{abstract}

%% file: Contents/1_introduction.tex
\section{Introduction}
\label{sec:introduction}

Widespread adoption of Machine Learning (ML) for critical tasks brought along the question of {\em trust}: Are ML models robust in making correct decisions when safety is at risk? Despite the significant advances in deep learning, near-human performance on several tasks did not translate to robustness in adversarial settings~\cite{szegedy2013intriguing,biggio2013evasion,athalye2017synthesizing,papernot2016transferability}. Several ML models, especially Deep Neural Networks (DNNs), are found susceptible to perturbed inputs resulting in potential safety implications~\cite{dreossi2018semantic, huang2011adversarial, janasosp}. While many defenses have been proposed to make ML models more robust, only a few have survived adaptive attack strategies~\cite{athalye2018obfuscated}. Among these defenses are certification~\cite{raghunathan2018certified} and its variants~\cite{cohen2019certified,lecuyer2018certified}, and adversarial training~\cite{madry-iclr2018}. The common theme across these defenses is to ensure that the model's output is stable within an $p$-norm ball around the input, where the $p$-norm is a crude proxy for human imperceptibility. 

However, these defenses suffer significant degradation in accuracy~\cite{tsipras2018there,jacobsen2019exploiting,cohen2019certified,lecuyer2018certified}. Recent results in literature have demonstrated that there are settings where robustness and accuracy cannot be simultaneously achieved~\cite{tsipras2018there,Bubeck:19,win-win}. Additionally, these defenses have largely been detached from the real world; they do not consider the semantic and contextual properties of the classification problem. Thus, the design of efficient and robust DNN classifiers remains an open research problem~\cite{carlini2019evaluating}. How can we defend against powerful adversaries and avoid the shortcomings of previous defenses? 


\begin{figure}[ht]%
    \centering
    \subfloat[Decision regions without invariant]{{\includegraphics[width=3.25cm]{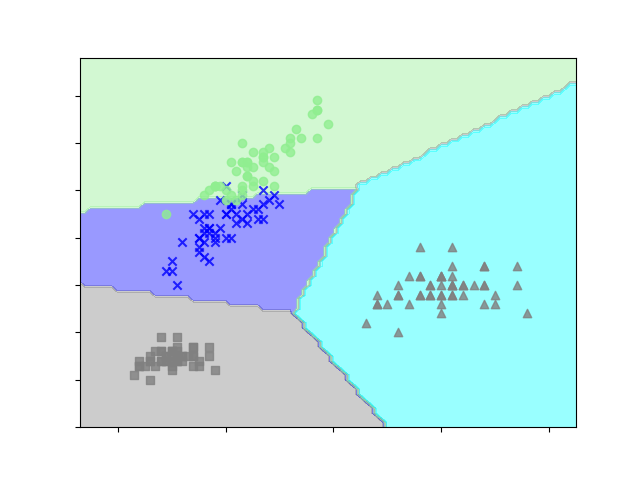} }}%
    \qquad
    \subfloat[Decision regions with invariant enforced]{{\includegraphics[width=3.25cm]{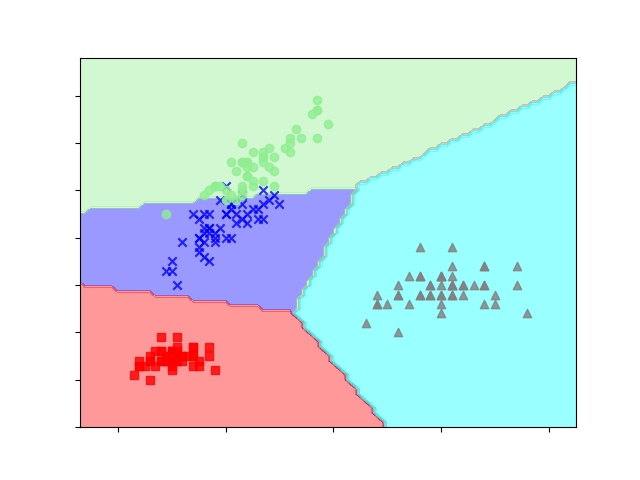} }}%
    \caption{\small Intuitive understanding of how invariances reduce adversarial mobility. In Figure 1(b), the bottom left red region is no longer accessible to the adversary.}%
    \label{fig:invariances}%
\vspace{-5mm}
\end{figure}

Fortunately, there is one constraint on the power of the adversary: \textit{the adversarial perturbation should not modify specific salient features which will change the perception/interpretation of the object.} Take the example of classifying a US road sign (Figure~\ref{fig:stop}). The shape of the sign is {\em intentionally} designed to be octagonal; few (if any) other signs have this particular shape. To attack such a sign, an attacker can no longer generate arbitrary perturbations; if the new shape deviates significantly from an octagon, then one can argue that the new object has deviated significantly from (and is no longer) the \texttt{Stop} sign. Thus, by forcing the adversary to satisfy these constraints (such as preserving the shape), we can increase the attack's cost relative to its resources, thereby satisfying Saltzer and Shroeder's ``work factor'' principle for the design of secure systems~\cite{saltzer1975protection}. 


These constraints posed by the defender can be modeled as {\em invariances}, which can be explicitly enforced in the classification procedure. When enforced, such invariances limit the adversary's attack space, consequently making the classifier more robust (refer Figure~\ref{fig:invariances}). Given an input domain $\mathcal{X}$, invariances (denoted $\mathcal{I}$) can be captured mathematically as a relation $\mathcal{I} \subseteq \mathcal{X} \times \mathcal{X}$. Observe that defense strategies that ensure that the label of a datapoint remains fixed in an $\varepsilon$-ball around it can also be captured in our framework \ie points $(x,x') \in \mathcal{X}$ belong to the invariant $\mathcal{I} = \{(x, x') : ||x - x'||_{p} \leq \varepsilon\}$ if and only if they have the same label.

To the best of our knowledge, there is no previous work that has investigated how to embed a set of invariances into a learning process. In this paper, we wish to understand how one can re-design the classification process to be {\em more robust} if they are given a set of pre-determined invariances. By doing so, we wish to understand if we can improve the trade-off between accuracy and robustness. Specifically, we discuss:

\vspace{1mm}
\noindent{\em Modelling domain knowledge as invariances:} We argue that classification problems in the wild can benefit from utilizing domain knowledge both in terms of improving its performance and improving robustness. Enforcing these invariances on an attacker introduces additional constraints on its search space. In particular, we prove that applying invariances affects results from literature that present statistical settings~\cite{tsipras2018there} (refer Appendix~\ref{tsipras}) and computational settings~\cite{win-win} (refer \S~\ref{sec:robustness}) where accuracy and robustness fail to co-exist.


\vspace{1mm}
\noindent{\em Data-driven approaches to obtain invariances:} We show that in scenarios where it is intractable to use domain expertise, clustering embeddings obtained from the training data can give sufficient insight to obtain invariances. Through simple experiments on the MNIST and CIFAR-10 datasets, we observe that the clusters generated by intermediary and ultimate layers (respectively) results in distinct label equivalences. We proceed to analyze if these equivalences can be used to design classification schemes with enhanced robustness (refer \S~\ref{sec:insights})

\vspace{1mm}
\noindent{\em Hierarchical Classification:} When enforced, the invariances split the output/label space of the classifier into equivalence classes, thereby partitioning the classification problem into smaller pieces (\eg refer Figure~\ref{fig:road-sign-setup}). Thus, we can design a hierarchical classification scheme (conceptually similar to a decision tree). At the decision nodes of the hierarchy, we have classifiers that enforce the invariances (by forcing inputs to label equivalence classes). At the leaves of the hierarchy, we have classifiers that predict within an equivalence class of labels (refer \S~\ref{sec:hierarchies} and \S~\ref{sec:general}). By ensuring that all of these constituents (leaves and intermediary classifiers) are robust, we are able to obtain a hierarchical classifier that is more robust than the sum of its parts.

\vspace{1mm}
\noindent{\em Robustness-Accuracy Improvement:} Employing the invariances within the hierarchical classifier provides robustness gains on two levels. At a high-level, a sequence of invariances limits the prediction to an equivalence class of labels. Equivalence classes limit the adversary's targeted attack capability; they can only target labels within the equivalence class. On a more fine-grained level, we show that reducing the number of labels improves the robustness certificate (defined in \S~\ref{certs}) of the classifier compared to the original classifier predicting within the full set of labels. More importantly, we show that these gains in robustness do not harm accuracy (refer \S~\ref{sec:robustness_analysis} and \S~\ref{sec:robustness_trade-off}).

As case studies of the real-world application of the invariances and the proposed hierarchical classifier, we study classification problems in two domains -- vision and audio. In the vision domain, we study the problem of road sign classification. Here, we demonstrate that {\em shape} invariances can be enforced/realized by using robust inputs. A realization of this approach is by predicting shape from robust \lidar point clouds as opposed to image pixels. From Figure~\ref{fig:lidar-teaser}, one can observe that the pixel-space inputs perturbed using adversarial patches from previous work~\cite{roadsigns17} is misclassified as a \texttt{SpeedLimit} sign (which is supposed to be circular), but the \lidar point cloud shows that the shape of the sign is still an octagon. In the audio domain, we study the problem of speaker identification. Here, we showcase another approach. Specifically, we show that {\em gender} invariances (through explicit gender prediction) can be enforced by using a classifier that is trained to be robust. Such a robust classifier will be able to perform predictions accurately despite using features that are not robust. 

Our results show that: 
\begin{enumerate}
\item Our approach is agnostic to the domain of classification. For both audio and vision tasks, we observe gains in robustness (sometimes a 100\% increase) and accuracy, suggesting that they may no longer need to be at odds.
\item The exact choice of implementation of different components of the hierarchy bears no impact on the correctness of our proposal. In \S~\ref{casestudy1}, we implement the root classifier of our hierarchy as a regular DNN trained to operate on robust features (obtained from a different input modality), and in \S~\ref{casestudy2}, we implement the root classifier as a smoothed classifier~\cite{cohen2019certified}. Both approaches result in a hierarchical classifier with increased robustness and accuracy.
\item By adding one additional invariant (location), we observe that there are significant gains in robustness for the vision task (refer \S~\ref{sec:morerobustfeatures}).
\end{enumerate}

%% file: Contents/2_background.tex
\section{Background}
\label{sec:background}

\subsection{Machine Learning Primer}
\label{sec:notation}

Consider a space $\mathcal{Z}$ of the form $\mathcal{X} \times \mathcal{Y}$ , where $\mathcal{X}$ is the input/sample space and $\mathcal{Y}$ is the output space. For example, in our case studies detailed in \S~\ref{expts}, the inputs are images of road signs or audio samples from speakers, and the outputs are the exact road sign or identities (respectively). Often, we assume $\mathcal{X}=\mathbb{R}^n$ and $\mathcal{Y} = \mathbb{R}^m$. Let $\mathcal{H}$ be a hypothesis space (\eg weights of a DNN). We assume a loss function $L:\mathcal{H} \times \mathcal{Z} \rightarrow \mathbb{R}$ that measures the {\em disagreement} of the hypothesis from the ground truth. The output of the learning algorithm is a classifier, which is a function $F$ which accepts an input $x \in \mathcal{X}$, and outputs $y \in \mathcal{Y}$. To emphasize that a classifier depends on a hypothesis $\theta \in \mathcal{H}$, which is output from the learning algorithm, we will denote it as $F_{\theta}$; if $\theta$ is clear from the context, we will sometimes simply write $F$. We denote by $F_i(x)$ as the probability that the input $x$ has the label $i \in [m]$, such that $0 \leq F_i(x) \leq 1$ and $ \sum_{i}{F_i(x)} = 1$.

\subsection{Threat Model}
\label{sec:threat}

Our threat model comprises of white-box adversaries that behave in a passive manner, and are capable of generating {\em human imperceptible} perturbations. Since human imperceptability is hard to measure, it is often approximated using the $p$-norm. Thus, the adversary's objective is to generate $p$-norm bounded perturbations (typical values for $p$ include 0, 1, 2, $\infty$); inputs modified using these perturbations are misclassified (\ie the label predicted by the classifier does not match the true label). The misclassifications could be targeted (\ie the target label which is to be output by the classifier is chosen by the adversary), or untargeted. Formally, an adversary wishes to solve the following optimization problem:
\begin{gather*}
\label{eq:original_attack}
\min \quad \|\delta\|_{p}, \quad \text{ s.t. } \\
\argmax F_i(x + \delta) = y^* \text{ (\textit{targeted}) } \\ 
\text{ or } \argmax F_i(x+\delta) \neq \argmax F_j(x) \text{ (\textit{untargeted}).} 
\end{gather*}

There exists abundant literature for generating such adversarial examples~\cite{madry-iclr2018,goodfellow2014explaining,moosavi2015deepfool,moosavi2017universal,papernot2017practical}. While adversaries are capable of generating perturbations that are not $p$-norm bounded~\cite{roadsigns17,kurakin2016adversarial}, such adversaries are out of the scope of this work; even in such a simple setting, progress has been limited. Additionally, the robustness measurements we make (as defined in \S~\ref{certs}) rely on the assumption that perturbations generated are $p$-norm bounded.

\subsection{Measuring Robustness}
\label{certs}

Several defense strategies have been devised to protect models/classifiers from adversarial examples~\cite{papernot2016distillation,samangouei2018defense,dhillon2018stochastic,meng2017magnet}; most of them have failed~\cite{athalye2018obfuscated}. Orthogonal research has focused on obtaining regions around an input where there provably exists no adversarial examples. This notion of {\em certified robustness} can be formalized by the $\varepsilon$-ball around $x$, defined as: $B_{p,\varepsilon}(x) = \{ z \in \mathcal{X} \mid \|z-x\|_p \leq \varepsilon \}$; the $\varepsilon$-ball of $x$ is a region in which no $p$-norm bounded adversarial example (for $x$) exists. The work of Lee \etal~\cite[\S2]{DBLP:journals/corr/abs-1906-04948} contains a detailed discussion of various certification methods and their shortcomings (primarily due to scalability). 

We center our discussion on the robustness certificate of Cohen \etal~\cite{cohen2019certified} for $2$-norm bounded perturbations. Their certificate computation scales for large DNNs, and several adversarial attacks focus on generating adversarial examples by bounding the $2$-norm, making their certificate relevant\footnote{Through the remainder of the paper, any discussion of robustness certificates refers to the one in~\cite{cohen2019certified}.}. The work of Lee \etal~\cite{DBLP:journals/corr/abs-1906-04948} generalizes the result in~\cite{cohen2019certified} for different distributions of noise and other norm bounds. 

\noindent{\bf Randomized Smoothing:} In a nutshell, Cohen \etal propose to transform any classifier to a smoothed classifier through the addition of isotropic Gaussian noise (denoted as $\eta$). We refer readers to~\cite[\S2,3,4]{cohen2019certified} for more details. They prove that the lower bound on the robustness certificate of each input $x$ for the smoothed classifier is given as:
\begin{equation}
\label{eq:robustnesscertificate}
    R \geq \frac{\sigma}{2}\left(\Phi^{-1}(\underline{p_A})-\Phi^{-1}(\overline{p_B})\right), 
\end{equation}
where (a) $\Phi^{-1}$ is the inverse of standard 
Gaussian CDF, (b) $\underline{p_A}$ is the lower bound of the probability of the top label, (c) $\overline{p_B}$ is the upper bound of the probability of the runner-up label, and (d) $\eta$ is drawn from  $\N(0,\sigma^2)$. They also devise an efficient algorithm to compute this certificate for large DNNs by empirically estimating the probability of each class's decision region  by running Monte Carlo simulations under the distribution $\N(x,\sigma^2)$. 

Other certified smoothing approaches have similar robustness bounds that are functions of the margin between the top and runner-up labels in the base classifiers~\cite{lecuyer2018certified, HeinA17}, so the trends in the results discussed in \S~\ref{expts} are independent of the actual certification methodology. 

%% file: Contents/3_invariants.tex
\section{The Building Blocks}

Through the remainder of the section, we discuss 
\begin{itemize}
\item how invariances can be used to restrict the adversary's freedom by introducing equivalence classes in the label space (\S~\ref{sec:invariances}). 
\item how label equivalences create a hierarchy, which provides an opportunity to introduce a hierarchical paradigm for classification (\S~\ref{sec:hierarchies}).
\item how using a combination of hierarchies and invariances with specific modifications to (a) the input, or (b) the model architecture, will result in robustness and accuracy no longer being at odds (\S~\ref{sec:robustness}).
 
\end{itemize}

\subsection{Hierarchies}
\label{sec:hierarchies}


In general, one can observe that if classification was performed in a hierarchical manner, various concepts could be verified/enforced at different levels of the hierarchy. Consider a simple example of a decision tree, where at each level, a local classification decision (based on a particular feature) occurs. In terms of understanding the relationship between hierarchies and test accuracy, one school of thought believes that knowledge distillation~\cite{hinton2015distilling,caruana2006model} using ensembles can drastically improve generalization performance. Another believes that concepts form hierarchies which can be exploited to improve accuracy~\cite{olah2017feature}. We investigate the latter. Several researchers have studied the notion of hierarchical classification~\cite{yan2015hd,deng2014large,salakhutdinov2012learning}; at a high-level, the early stages of the hierarchical classifier performs classification based on high-level concepts that exist between various datapoints (such as shape, or color), and the later stages of the hierarchical classifier performs more fine grained classification. The intuition for using a hierarchy stems from earlier observations where DNNs extract high-level features (of the input images) in the earlier layers~\cite{zeiler2014visualizing}; researchers believed explicitly enforcing this would result in better performance~\cite{HD-CNN,category_structure,treepriors_transferlearning}. Prior work contains abundant literature on defining architectures to optimize for improving the test performance of machine learning models. In the context of DNNs, neural architecture search is one such emerging area~\cite{zoph2016neural}. 

\begin{figure}
  \centering
  \includegraphics[width=0.5\linewidth]{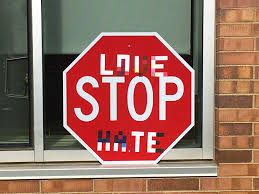}
  \caption{\small A US \texttt{Stop} sign with stickers. Figure obtained from~\cite{roadsigns17}.}
  \label{fig:stop}
\end{figure}

Intuitively, this process is inspired by how humans {\em may} perform classification~\cite{national2008human,hunt1962nature,dunsmoor2015categories,lopez1992development}. Consider the case where a human is tasked with identifying a road sign. One of the first observations the human makes with regards to the road sign is its shape, or its color, or its geographic location. These observations provide the human with priors, and consequently enable the human to perform more accurate classification (based on these priors). In essence, these priors encode abundant context for the humans, enabling better classification performance. For example, if the human was asked to identify the sign in Figure~\ref{fig:stop}, the human may first reduce the potential candidates for the sign based on the {\em shape} of the sign, and then make a final classification among different signs of the {\em same shape.} Geirhos \etal~\cite{geirhos2017comparing,geirhos2018generalisation} make a similar observation; they observe that humans are more robust to changes in the input (such as textural or structural changes) potentially due to the (hierarchical) manner in which they perform classification. DNNs operate in a similar manner, by extracting filters that behave similar to human perception~\cite{zeiler2014visualizing}.



\subsection{Invariances}
\label{sec:invariances}

From the discussion in \S~\ref{sec:hierarchies} (and the related work in \S~\ref{sec:related_work}), we observe that encoding priors about the classification task can help a model learn better. Our contribution is to demonstrate how encoding specific information (regarding the classification task, input features, network architectures etc.) as invariances can help increase the robustness. The inevitable existence of adversarial examples suggest that DNNs are susceptible to very subtle changes in the input~\cite{DBLP:journals/corr/abs-1809-02104}. However, humans are (significantly) more robust to such changes\footnote{Earlier work suggests that humans are fooled by certain type of inputs~\cite{elsayed2018adversarial}.}, and can even identify these perturbations~\cite{zhou2019humans}. In the case of road sign classification, the work by Eykholt \etal~\cite{roadsigns17} introduces realizable perturbations (those which are larger than the $p$-norm bounded perturbations considered earlier) that fool DNNs, but humans are still able to classify these signs correctly. We believe that this is due to the priors that humans have regarding classification tasks (\eg in Germany, hexagonal road signs can only be \texttt{Stop} signs, yellow signs can only be \texttt{SpeedLimit} signs etc.). To further motivate our discussion, Geirhos \etal observe that DNNs are more susceptible to textural changes~\cite{geirhos2018imagenet}, and {\em explicitly enforcing} bias towards shapes can improve robustness. 

We believe that this observation is more general; by enforcing such biases which we collectively call invariants/invariances, one can improve DNN robustness. Mathematically, we denote an {\em invariant} $\mathcal{I} \subseteq \mathcal{X} \times \mathcal{X}$. For example, let $S : \mathcal{X} \rightarrow \mathcal{Z}$ denote the function that predicts the shape of input $x \in \mathcal{X}$. Thus, the {\em shape} invariant can be formalized as follows: $\{(x, \tilde{x}) \in \mathcal{I} \text{ iff } S(x) = S(\tilde{x}) \}$. 

Several questions arise, first of which being {\em how does one identify the invariances related to a particular task?} This is a challenging problem. In our evaluation in \S~\ref{sec:insights} and \S~\ref{expts}, we show two different approaches: (a) by measuring similarity between datapoints' embeddings in different feature representations, and (b) domain expertise related to the classification task. These approaches broadly encompass two different types of invariances: human un-interpretable and human interpretable. While the exact nature of the invariant is not important for methodology, interpretability/explainability may help in debugging and understanding model failures. We discuss this in detail in \S~\ref{sec:discussion}. 

We believe that both approaches have merit; while DNNs have displayed the capability to generalize, they are often used for very specific tasks in the real-world, motivating how domain expertise can be useful. Recent work in adversarial training~\cite{new_madry_2019} suggests that human imperceptible features are potentially useful in improving robustness, and these features can be extracted by using axiomatic attribution methods~\cite{sundararajan2017axiomatic}, or observing intermediary representations~\cite{papernot2018deep}. 

Note that the existence of invariances induces a hierarchy in the labels. In the road sign classification example, assuming that shape is invariant, road signs can be grouped based on their shape, and then within each group, partitioned in a more fine-grained manner based on their individual label (refer to Figure~\ref{fig:road-sign-setup}). Thus, we observe that invariances partition the label space into equivalence classes. The existence of equivalence classes is analogous to the existence of hierarchies, and this ties well with the hierarchical classification paradigm discussed earlier.

\noindent{\bf Takeaway 1:} Invariances are task specific, and can be obtained using domain knowledge in some cases, and by analyzing properties of the dataset in others.

The next question is {\em how to encode such invariances?} Before we discuss this further, it is important to note that invariances are a property of the data distribution that we analyze and the model that is being used to learn/predict with this data. Changes to either will result in datapoints (or the embeddings they generate) violating the invariances. 

The easiest way to encode invariances is to add additional features that are directly correlated to the invariances. For example, one could add an additional feature to an image of a road sign that is suggestive of its shape. However, assuming that the adversary has no access to this additional feature is too strong. An alternative approach is to encode invariance information in the standard features used for classification; now these features have two purposes: they contain useful information regarding the invariance, and they contain useful information about how the input can be classified. However, the existence of adversarial examples suggests that {\em all} input features may not be useful in encoding information related to the invariances (as some input features are very susceptible to adversarial perturbations). To this end, Madry \etal~\cite{madry-iclr2018} define {\em robust features} as those which are hard to adversarially perturb, but are highly correlated with the correct classification output. Thus, one can utilize these robust features for invariance enforcement\footnote{It is important to note that robust features, by themselves, are not sufficient for performant classification~\cite{madry-iclr2018}}. For example, one could identify robust features in a \texttt{Stop} sign image and use them to verify if it is indeed hexagonal. Alternatively, one could use domain knowledge to obtain robust features (by strategically modifying the inputs to obtain a different set of input features). In our evaluation in \S~\ref{sec:insights}, we highlight how robust features are useful for invariance enforcement. In our case study in \S~\ref{casestudy1}, we show how inputs from a different modality can be robust, and can be used for invariance enforcement. However, in the absence of robust features, or a different set of robust input features, invariance encoding happens in the vulnerable features. While this is not as accurate as the other two approaches, we discuss how this can be useful next.

\noindent{\bf Takeaway 2:} Ideally, invariances can be encoded using robust (input) features. In the absence of such robust features, invariances can be encoded in the vulnerable features as well.

The final question is {\em how to verify if the invariance has been preserved?} If robust features are used for encoding invariance information, we can use an off-the-shelf classification model to check if the invariance has been preserved by observing the outcome of the classification \ie if the input falls into the correct equivalence class. For example, if the robust features obtained from an adversarially perturbed \texttt{Stop} sign enables it to be classified into the {\em hexagonal} equivalence class, it implies that the {\em shape} invariance has been preserved. In the absence of robust features, invariance encoding occurs in the vulnerable features; one could train a model to be robust to check if the invariance is preserved in the features. We describe a realization of this approach in \S~\ref{sec:insights}.

\noindent{\bf Takeaway 3:} Since the invariances partition the label space into equivalence classes, verifying if the invariance is preserved is as simple as checking if the input (benign or adversarially perturbed) falls into the correct equivalence class.

We would like to stress that the connection between invariances and stability (of learning) is not new; Arjovsky \etal~\cite{arjovsky2019invariant} suggests that invariant properties (or stable properties as referred in their work) are essential in improving generalization. As they note, to improve robustness, one must detect (and enforce) these stable properties to avoid spurious correlations that exist in data. Similarly, Tong \etal~\cite{tong2017improving} state that it is hard to simultaneously modify, what they term, conserved features and preserve functionality (in the case of malware). 

\subsection{Robustness}
\label{sec:robustness}

We can see how invariances can potentially restrict the adversary's attack space. We formalize how these restrictions result in increased robustness by highlighting the construction of Degwekar \etal~\cite{win-win} based on Pseudo-Random Functions (PRFs). This construction shows a scenario where a robust classifier exists~\cite{Bubeck:19} but is hard to find in polynomial time, but under the existence of specific invariances, it becomes easy to learn a robust classifier.

\noindent{\em Construction:} Let $\mathcal{B} = \{0,1\}$, and $F_k : \mathcal{B}^n \rightarrow \mathcal{B}$ be a PRF with a secret key $k$. Let $(Encode,Decode)$ be an error-correcting code (ECC), such as those in~\cite{GI01}, where $Encode$ encodes a message and $Decode$ decodes a message, and $Decode$ can tolerate a certain number of errors in its input. Consider the following two distributions:
\begin{eqnarray*}
D_0 & = & \{0,Encode\left(x,F_k(x)\right)\} \\
D_1 & = & \{1,Encode\left(x,1-F_k (x)\right)\},
\end{eqnarray*}
where $x$ is drawn uniformly from $\mathcal{B}^n$, and `,' denotes concatenation. Given these distributions, there exists a classifier to distinguish between the two because the first bit always indicates which distribution ($D_0$ or $D_1$) it belongs to. This classifier has perfect natural accuracy and is easy to learn~\cite{Bubeck:19}. This classifier is also robust (under certain assumptions which we explain next). Due to the properties of the ECC, $R$ can tolerate a constant fraction of the errors among {\em all but the first bit}. If any attacker flips the first bit, a robust (or any) classifier is hard to learn if $R$ is unaware of $k$. However, if $R$ has access to the secret key $k$, then given an $\tilde{x} \sim D_b$ where $b \in \mathcal{B}$, $R$ first executes $Decode(\tilde{x}_{1:})$ (where $\tilde{x}_{i:}$ denotes the the last $n-i$ bits of $\tilde{x}$ \ie $\tilde{x} = \tilde{x}_{0:}$) and obtains $x$. Then, the classifier can then check the last bit $\tilde{x}_{n-1}$ to see whether it is $F_k(x)$ or $1-F_k(x)$. Without knowing the key $k$, the verification described earlier would not always be accurate. This result follows from the fact that $F_k(\cdot)$ is a PRF (\ie essentially a probabilistic polynomial time adversary cannot distinguish between $F_k (x)$ and a random bit)~\cite{win-win}. This setting demonstrates a situation where {\it a robust classifier exists but cannot be found in polynomial time (without knowledge of $k$).}

However, if the following invariant $\mathcal{I} = \{(x,\tilde{x}) | x_0 = \tilde{x}_0\}$ (where $x_i$ denotes the $(i+1)^{th}$ bit of $x$) is enforced, then the hardness result above is trivially negated. Thus, it is clear how invariances can enable robustness. 

Intuitively, invariances help by limiting the attacker's actions (refer Figure~\ref{fig:invariances}). Recall from our earlier discussion that these invariances also induce label equivalences. For the remainder of this work, we focus on those invariances that produce {\em disjoint label equivalence classes}; this property makes our analysis easier (but has no bearing on our approach).

As stated earlier, we measure robustness through the certificate provided by the work of Cohen \etal~\cite{cohen2019certified}\footnote{However, we do show the generality of our approach using adversarial training~\cite{madry-iclr2018} and standard datasets (such as MNIST and CIFAR-10) in \S~\ref{sec:insights}}. Recall that the certificate is proportional to the margin between the probability estimates for the {\em top candidate} for the label and the {\em runner-up} \ie $R \propto \Phi^{-1}(\underline{p_A})-\Phi^{-1}(\overline{p_B})$. Thus, it is clear that one can increase the certificate by increasing this margin. In \S~\ref{casestudy1}, we provide detailed arguments and constructions on how we can obtain an increased margin. 

In principle, one could envision a scenario where there exist multiple invariances (at different levels of the hierarchy), resulting in (disjoint) label equivalence classes of size 1 (at the leaves of the hierarchy). In such a scenario, the robustness certificate for each equivalence class is $\infty$ (even perturbations with large $p$-norms will not cause misclassifications).

%% file: Contents/5_general_framework.tex
\section{A General Framework}
\label{sec:general}

In this section, we provide a generalization of the construction required to construct DNNs for increased robustness (\S~\ref{sec:hier_construction}). We show how robustness can be enhanced by increasing the depth of the hierarchy with many invariances being enforced (\S~\ref{sec:robustness_analysis}), and the guarantees this provides (\S~\ref{sec:robustness_trade-off}). 

\subsection{Components of a Hierarchical Classifier}
\label{sec:hier_construction}


\begin{figure}
  \centering
  \includegraphics[width=\linewidth]{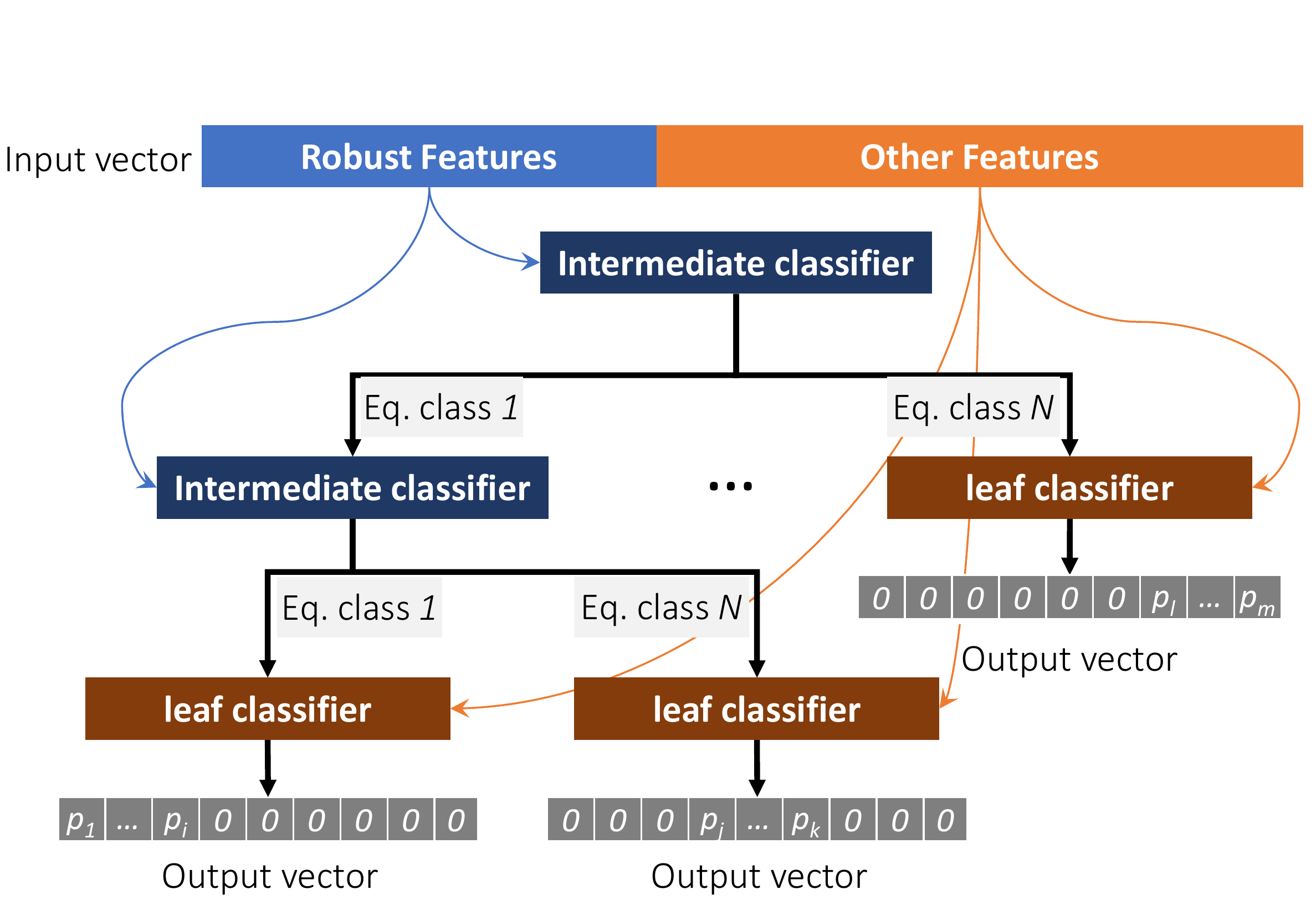}
  \caption{\small High-level description of the hierarchical classifier. The thin arrows highlight data flows while thick arrows indicate decision paths. The original set of labels is $\{1, \ldots, m\}$. Each intermediate classifier splits the label set further. Each leaf classifier predicts within a reduced set of labels. For example, the left-most classifier assigns each label within $\{1, \ldots, i\}$ a probability value while assigning the other labels $\{i+1, \ldots, m\}$ a probability of 0. }
  \label{fig:high-level}
  \vspace{-5mm}
\end{figure}

The hierarchical classifier is a sequence of classifiers in a hierarchy, where each classifier at a level of the hierarchy operates over a subset of the feature space. Observe that we make no assumptions about the nature of these feature subsets, or the nature of the different classifiers in the hierarchy. Just like the conventional flat classifier, the hierarchical classifier is a function $F: \mathcal{X} \rightarrow \mathcal{Y}$ However, all levels other than the leaf classifiers are useful in preserving the invariances; only the leaf classifier predicts labels. Figure~\ref{fig:high-level} show the high-level structure of the hierarchical classifier, including the input features, classifiers, and output vectors. We now describe these components.


\vspace{1mm}
\noindent{\em 1. Intermediate Classifiers:} As explained earlier, the objective of the intermediary classifiers is to ensure that the invariances are preserved. To do so, intermediary classifiers are either (a) robustly trained~\cite{madry-iclr2018,cohen2019certified} such that they can forward the input to the right equivalence class induced by the invariances, or (b) accept robust features (as defined in \S~\ref{sec:invariances}) as inputs to ensure such forwarding. Since the classifiers are either robust themselves, or operate on robust inputs, they are hard to attack. Thus, an intermediate classifier $F_{intermediate}: \mathcal{X} \rightarrow \mathcal{K}$, where $\mathcal{K}$ denotes the equivalence regions in the label space $\mathcal{Y}$ \ie $\mathcal{K}=\{i: \mathcal{Y}_i \subset \mathcal{Y}\}$. We do not assume that our hierarchy is balanced, and make no assumptions about the number or type (linear vs. non-linear models) of such intermediary classifiers\footnote{There is a correspondence between an invariance and an intermediate classifier; each intermediate classifier is required to enforce a particular invariance. However, this mapping is not necessarily bijective.} (refer Figure~\ref{fig:high-level}). It is important to note the ordering of these intermediary classifiers has a direct impact on accuracy, but not on robustness (which is only dependent on the label equivalences, which in turn depends on the invariances causing them); independent of the ordering of these intermediary classifiers, the subset of the label space in each leaf node is the same \ie the order of the intermediary classifiers is commutative. Since the robustness radius is a function of the labels predicted by the leaf, the earlier ordering does not impact the robustness calculation. Determining this exact ordering to maximize accuracy is left as future work. 

\vspace{1mm}
\noindent{\em 2. Leaf Classifiers:} A leaf classifier makes the final classification. More formally, for each equivalence class $i \in \mathcal{K}$, there exists a leaf classifier $F_{i}: \mathcal{X} \rightarrow \mathcal{Y}_i$. Observe that intermediate classifiers predict the equivalence classes, while leaf classifiers predict {\em labels} within an equivalence class. To train such a classifier, only the samples which have labels within $\mathcal{Y}_i$ are needed. Thus, the overall inference procedure (from root to leaf) is very similar to the decision tree; each intermediate classifier chooses the next one to be invoked till the inference reaches a leaf classifier. Only one leaf classifier is invoked for the entire input. Intermediate classifiers are trained to be robust so that they can correctly predict between different equivalence classes. Leaf classifiers are also trained in a robust manner so that they can correctly predict {\em within} a particular equivalence class. However, training different leaf classifiers (depending on the exact equivalence class) is a computationally expensive procedure. While the hierarchical classification paradigm makes no assumptions about the exact nature of the classifier used at the leaf, we make certain assumptions for our evaluation. We assume that all leaf classifiers have the same model (\eg DNN). Thus, we can utilize two strategies -- {\em retraining} and {\em renormalization} -- to obtain leaf classifiers. In \S~\ref{retrainvsrenormalize}, we discuss both these approaches in detail, as well as their pros and cons.

\subsection{Robustness Analysis}
\label{sec:robustness_analysis}

The hierarchical classifier forces the attacker into an equivalence class of labels and limits its {\em targeted attack capability}; an attacker cannot move the input outside an equivalence class. The leaf classifier, predicting within reduced label set, improves the robustness certificate by making the classifier stable within a larger ball around the input, limiting the attacker's capability within the equivalence class. This robustness property is subtle; it arises from the observation that reducing the labels can potentially widen the margin between the best prediction and the runner-up prediction. Below, we show that this property holds for any general classifier. 

Let $\alpha: \R^n \rightarrow \Delta (\mathcal{Y})$, where $\Delta(\mathcal{Y})$ is the set of all distributions over the labels $\mathcal{Y}$. For ease of notation, $\alpha (x)_l$ (for $l \in {\mathcal Y}$) denotes the
probability corresponding to $l$ in $\alpha (x)$. Fix $c \in \mathcal{Y}$, and for $L \subseteq \mathcal{Y}$, we define $H_{c,\alpha} (x,L)$ as follows:

\[
  H_{c,\alpha} (x,L) = \alpha (x)_c - \max_{l \in L \; \wedge \; l \not= c} \alpha(x)_l
\]

\begin{figure}
  \centering
  \includegraphics[width=0.8\linewidth]{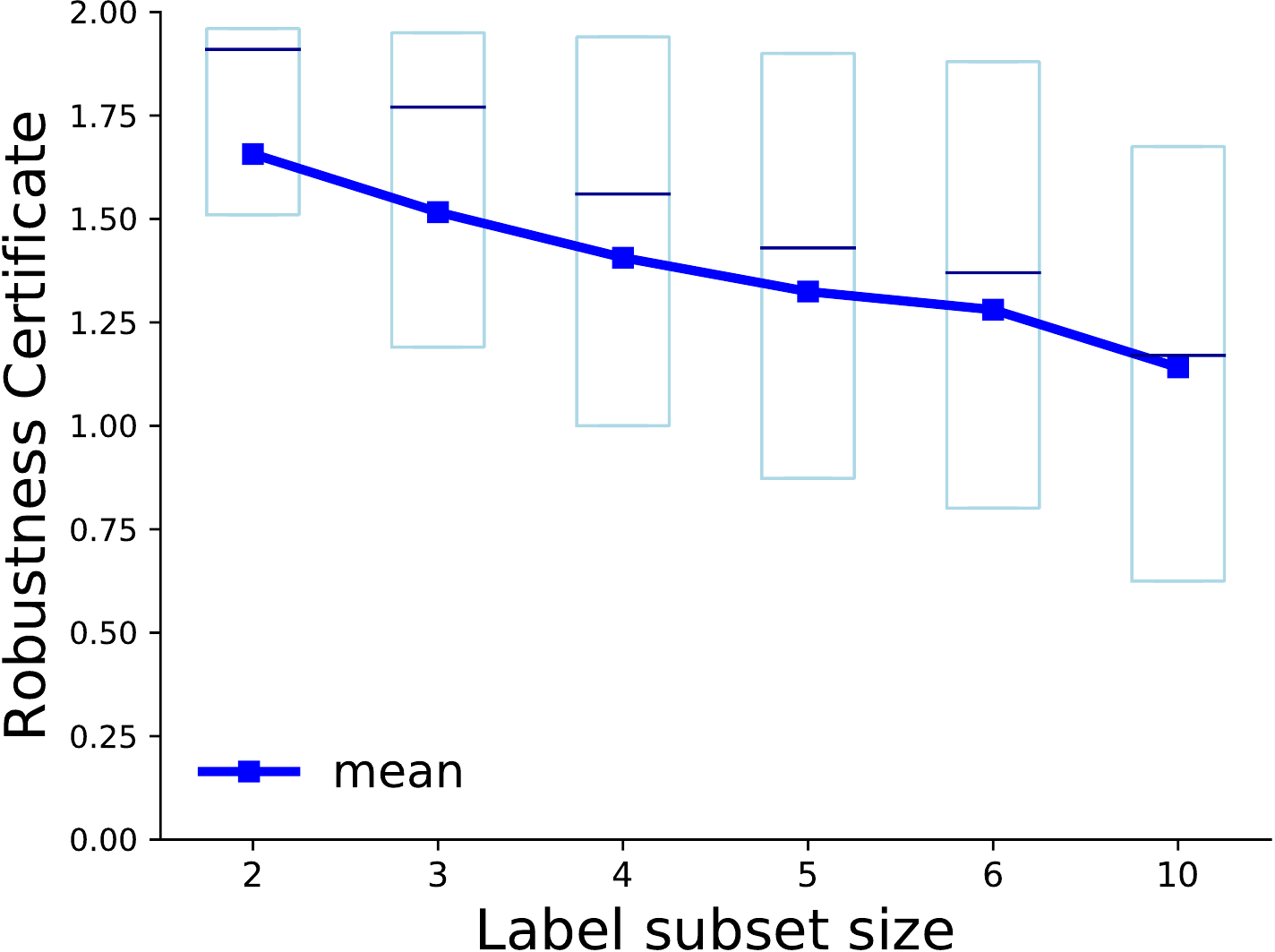}
  \caption{\small Box plot containing the robustness radius as function of the size of label sets in CIFAR-10, using the randomized smoothing approach of Cohen \etal~\cite{cohen2019certified}. We generate all label subsets of a particular size, and plot the robustness certificate values. Observe that as the subset size decreases, the mean robustness radius increases.}
  \label{fig:kolter-toy-example}
  \vspace{-5mm}
\end{figure}

It is easy to see that if $L_1 \subseteq L_2$, then
$H_{c,\alpha} (x,L_2) \leq H_{c,\alpha} (x,L_1)$, or $H_{c,\alpha}$ is anti-monotonic in the second parameter. Recall that $H_{c,\alpha}$ is similar to ``hinge loss''. If we instantiate $\alpha$ by the output of the softmax layer and use the argument of Hein \etal~\cite{HeinA17} for any classifier, we can immediately see that robustness radius increases as the set of possible labels is decreased. A similar argument can be used for the smoothing approaches~\cite{cohen2019certified,lecuyer2018certified}. For example,  Figure~\ref{fig:kolter-toy-example} shows the robustness radius for label subsets of different sizes from CIFAR-10.
The robustness radius is computed using the randomized smoothing approach of Cohen \etal~\cite{cohen2019certified}. It is evident from Figure~\ref{fig:kolter-toy-example} that as the subset size decreases, the average robustness certificate per sample increases.

\subsection{Robustness Guarantees}
\label{sec:robustness_trade-off}

The robustness guarantees of the entire hierarchical classifier depends on each individual classifier (the intermediate classifiers and the leaf classifiers, collectively called internal classifiers). Given a set of internal classifiers, the attacker needs to attack only one of them to change the classification output in the untargeted attack scenario (or attack a specific set of internal classifiers in the targeted attack scenario). Then, the robustness guarantee of the hierarchical classifier is the minimum of the guarantees of its constituent classifiers. 

To see why the robustness guarantee of the hierarchical classifier is the minimum of the guarantees of the composing classifiers, consider the simple case of three classifiers: $f^1$, $f^2$, and $f^3$ which form a larger classifier $F$. The hierarchy is such that $f^1$ is a binary classifier deciding between passing the input to $f^2$ or $f^3$, which are the leaf classifiers. A white-box adversary aims to attack the larger classifier $F$ as usual: 
 \begin{equation}
 min \|\delta\|_p \, \text{ s.t. } \, \argmax F(x+\delta) \neq \argmax F(x)
\end{equation}

Using the internal knowledge of the classifier, the adversary's objective can be restated as: 
 \begin{equation}
 \begin{gathered}
min \|\delta\|_p \, \text{ s.t. }\\
\argmax f^1(x+\delta) \neq \argmax f^1(x) \\
\text{\bf or } (\argmax f^1(x+\delta) = \argmax f^1(x) \\
\wedge \argmax f^2(x+\delta) \neq \argmax F_j(x)) \\
\text{\bf or } (\argmax f^1_i(x+\delta) = \argmax f^1(x) \\
\wedge \argmax f^3(x+\delta) \neq \argmax F(x))
\end{gathered}
\end{equation}

Since only one of the constraints has to be satisfied, the problem can broken down into smaller subproblems: 
\[min \|\delta\|_p = \min\left( \|\delta_1\|_p, \|\delta_2\|_p, \|\delta_3\|_p \right),\] where:
 \begin{equation}
 \begin{gathered}
 \|\delta_1\|_p = min \|\delta\|_p \, \text{ s.t. } \argmax f^1(x+\delta) \neq \argmax f^1(x) \\
\|\delta_2\|_p = min \|\delta\|_p \, \text{ s.t. } (\argmax f^1(x+\delta) = \argmax f^1(x) \\
\wedge \argmax f^2(x+\delta) \neq \argmax F(x)) \\
\|\delta_3\|_p = min \|\delta\|_p \, \text{ s.t. }  (\argmax f^1(x+\delta) = \argmax f^1(x) \\
\wedge \argmax f^3(x+\delta) \neq \argmax F(x))
\end{gathered}
\end{equation}

We can take the lower bound $\|\delta_2\|_p$ and $\|\delta_3\|_p$ by solving the less constrained problem of:
 \begin{equation}
 \begin{gathered}
\|\delta_2\|_p \geq \|\delta'_2\|_p = min \|\delta\|_p \, \text{ s.t. }  \\
\argmax f^2(x+\delta) \neq \argmax F(x) \\
\|\delta_3\|_p \geq \|\delta'_3\|_p = min \|\delta\|_p \, \text{ s.t. }  \\
\argmax f^3(x+\delta) \neq \argmax F(x).
\end{gathered}
\end{equation}

Finally, the lower bound of the needed perturbation to attack $F$ is the minimum of the perturbations needed to attack each network individually. In particular,  $min \|\delta\|_p \geq \min\left( \|\delta_1\|_p, \|\delta'_2\|_p, \|\delta'_3\|_p \right)$. This example can be generalized for multiple and non-binary intermediate classifiers. 

%% file: Contents/4_motivation.tex
\section{Early Insights}
\label{sec:insights}

Using two datasets (CIFAR-10~\cite{krizhevsky2014cifar} and MNIST~\cite{lecun2010mnist}), we show how hierarchies used to enforce invariances can provide robustness gains using other robustness techniques such as adversarial training~\cite{madry-iclr2018}. We also display a proof-of-concept approach to suggest that invariances are easy to obtain in a data-guided manner. The setup for our experiment is described in Appendix~\ref{app:setup}.

\subsection{Obtaining Invariances}
\label{sec:obtaining}

\begin{algorithm}
\caption{Data-driven equivalence class generation}
\begin{algorithmic}[1]

    \State Obtain embeddings \texttt{C} of all points in the dataset by observing the output of a DNN \texttt{F} at layer \texttt{n}. 
    \State \texttt{clusters} $\leftarrow$ \texttt{ClusterAlg}(\texttt{C})
    \State Verify number of centroids $\lvert$\texttt{clusters}$\lvert$ = $k$.
    \State Verify that each cluster in \texttt{clusters} is separated.
    \For{i = 1 $\cdots$ k}
        \State \texttt{cluster} $\leftarrow$ \texttt{clusters[i]}
        \State $l_i \leftarrow$ {\em unique} labels of embeddings in \texttt{cluster}
        \State \texttt{L[i]} = \texttt{L[i]} $\cup$ $l_i$ 
    \EndFor
    \If{$\cap_i$\texttt{L[i]}$=\phi$}
        \State Each \texttt{L[i]} denotes a label equivalence class.
    \Else
        \State Repartition labels in $\{$\texttt{L[i]}$\}_{i=1}^k$ s.t. $\cap_i$\texttt{L[i]}$=\phi$ 
    \EndIf
    \State Construct classifier \texttt{F}$_{intermediate}$ to classify inputs to an equivalence class in $\{1,\cdots,k \}$ such that the corresponding label equivalence classes are \texttt{L[1]}$\cdots$\texttt{L[k]}  
\end{algorithmic}
\end{algorithm}

\vspace{1mm}
\noindent{\bf MNIST:} To obtain the invariances, we first trained a small CNN model\footnote{3 convolutional layers followed by a FC layer and a softmax layer} on the MNIST dataset. We gather embeddings\footnote{In the context of neural networks, embeddings are those low-dimensional, continuous vector representations of discrete variables that are learned by the network.} from the intermediary layers, and cluster them using k-means clustering (similar to the approach proposed by Papernot \etal~\cite{papernot2018deep}). We observe that for $k=2$, the digits 2,3,0,6,8,5 form a cluster, while the others form another cluster.

\noindent{\bf CIFAR-10:} To obtain the invariances, we first adversarially train a CNN\footnote{Wide Resnet with 34 layers} on the CIFAR-10 dataset. We then observe the output of an adversarially trained CNN to produce a confusion matrix (refer Figure~\ref{fig:confusionmatrix1} in the Appendix). From this matrix, one can clearly observe two equivalence classes partitioning the label space -- non-living objects and living objects. Note that this invariance is also human understandable.

\vspace{1mm}
Note that the clustering could potentially be different if the outputs of a different layer was observed. However, it was not the case for our experiments. Additionally, this experiment just serves as a proof-of-concept. We believe that with such forms of intermediary clustering, one can automate the approach of creating such equivalence classes in a way as to avoid having domain specific knowledge. 

\subsection{Methodology}
\label{sec:method}

We observe that for both datasets, we were able to obtain an invariant that partitions the label space into two equivalence classes. Thus, to enforce such an invariant, we train a {\em root} classifier to classify inputs into one of these two equivalence classes. To make sure the root is robust to adversarial examples, we train the root classifier using adversarial training~\cite{madry-iclr2018} (to minimize the $\ell_\infty$-norm of the generated perturbation). For the CIFAR-10 dataset, the root classifier has a natural accuracy of 96.57\% and adversarial accuracy of 84.25\%. For the MNIST dataset, the root classifier has a natural accuracy of 98.51\% and adversarial accuracy of 93.68\%. We detail all parameters used for our experiments in Appendix~\ref{app:insights}. Thus, each hierarchy comprises of one root and two leaf classifiers.

\subsection{Results}
\label{prelim_results}

\begin{table}[h]
\small
\begin{center}
  \begin{tabular}{p{1.5cm}  p{1.5cm}  p{1.5cm}  p{1.5cm}}
    \toprule
    \text{\footnotesize Model} & \text{\footnotesize Natural (\%)} & \text{\footnotesize Adv (\%)} & \text{\footnotesize Budget (\%)} \\
    \midrule
    {\footnotesize Baseline} & {\footnotesize 98.79\%} & {\footnotesize 87.13\%} & {\footnotesize 87.13\%}\\
    {\footnotesize Hierarchy} & {\footnotesize 97.51\%} & {\footnotesize 84.72\%} & {\footnotesize 90.42\%}\\
    \bottomrule
    \end{tabular}
\caption{\small For the MNIST dataset and the invariances described in \S~\ref{sec:obtaining}), observe that adversarial accuracy, denoted Adv (\%), decreases in the hierarchical classifier compared to the baseline. The legends are (a) {\bf Natural (\%)}: natural accuracy, (b) {\bf Adv (\%):} adversarial accuracy from PGD attack where an adversary attacks the root and 2 leaf classifiers, and (c) {\bf Budget (\%):} restricted adversarial accuracy using PGD where an adversary is only able to attack one classifier chosen from the hierarchy (in our example: the leaf classifier for the 6 classes).}
\label{table:mnist_adv}
\end{center}
\vspace{-5mm}
\end{table}

\noindent{\bf MNIST:} For the given invariances, we observe a {\em decrease} in adversarial accuracy by 2.41 percentage points in our hierarchical classification (refer Table~\ref{table:mnist_adv}). This suggests (a) that the invariant obtained using the data-driven methodology is not the best for increasing robustness, or (b) there are some classes of invariances that potentially cause more harm than good (we discuss this in \S~\ref{sec:discussion}). This further motivates our other approach of obtaining invariances using domain knowledge. 


\begin{table}[h]
\small
\begin{center}
  \begin{tabular}{p{1.5cm}  p{1.5cm}  p{1.5cm}  p{1.5cm}}
    \toprule
    \text{\footnotesize Model} & \text{\footnotesize Natural (\%)} & \text{\footnotesize Adv (\%)} & \text{\footnotesize Budget (\%)} \\
    \midrule
    {\footnotesize Benchmark} & {\footnotesize 83.99\%} & {\footnotesize 44.72\%} & {\footnotesize 44.72\%}\\
    {\footnotesize Baseline} & {\footnotesize 83.68\%} & {\footnotesize 38.77\%} & {\footnotesize 38.77\%}\\
    {\footnotesize Hierarchy} & {\footnotesize 85.37\%} & {\footnotesize 45.46\%} & {\footnotesize 59.11\%}\\
    \bottomrule
    \end{tabular}
\caption{\small For the CIFAR-10 dataset and the invariances described in \S~\ref{sec:obtaining}), observe that adversarial accuracy, denoted Adv (\%), increases in the hierarchical classifier compared to the baseline and the benchmark published in YOPO\cite{yopo}. The legends are (a) {\bf Natural (\%)}: natural accuracy, (b) {\bf Adv (\%):} adversarial accuracy from PGD attack where an adversary attacks the root and 2 leaf classifiers, and (c) {\bf Budget (\%):} restricted adversarial accuracy using PGD where an adversary is only able to attack one classifier chosen from the hierarchy (in our example: the leaf classifier for the Living objects equivalence class).}
\label{table:cifar_adv}
\end{center}
\vspace{-5mm}
\end{table}


\noindent{\bf CIFAR-10:} With our hierarchical classification scheme, we achieve an increase in adversarial accuracy by 6.69 percentage points when compared to a baseline classifier. To measure the adversarial accuracy\footnote{Adversarial accuracy for our hierarchy is defined as $\sum \frac{x}{X}$ where x is the number of correct predictions for each leaf classifier and X is the total number of samples in the test set.}, we imagine the following 2 white-box attack scenarios:
\begin{enumerate}
    \item {\em Worst-Case}: An unrestricted adversary with access to each of the models in the hierarchy.
    \item {\em Best-Case}: A restricted adversary which is limited to attacking only one of the classifiers in the hierarchy.
\end{enumerate}
To simulate the worst case adversary, we begin by attacking the root classifier (with $\ell_\infty$-norm perturbed adversarial examples). For those inputs which did not cause misclassification, we generate adversarial examples for the leaves. While this attack is stronger (as the adversary has multiple chances at creating the adversarial example), this attack is also computationally heavier. In the case of a budgeted adversary, the attacker would attack the classifier which results in the largest drop in accuracy; in our evaluation, this happened to be the leaf classifier related to classifying animate objects. Observe that in all scenarios described, the hierarchy produces increased gains (refer Table~\ref{table:cifar_adv}). We compute the robustness certificates for these networks (by tweaking some components of the training process) and report the results in Appendix~\ref{app:cifar}. We also observe an increase in robustness as measured by the certificate.

%% file: Contents/6_experiments.tex
\section{Case Studies}
\label{expts}


In \S~\ref{sec:insights}, we witnessed improvements for robustness in a toy setting. In this section, we investigate our approach in more realistic settings. Specifically, we wish to obtain answers for the following questions:

\begin{enumerate}
    \item Is the proposed approach valid (and useful) only for tasks pertaining to vision?
    \item Does the choice of implementation of the root and intermediary classifiers (either by using robust features, or training the classifier to be robust) impact our approach in practice?
    \item Does the number of invariances used impact the gains in robustness?
\end{enumerate}

To answer these questions, we implement a hierarchical classifier for classification tasks in two domains: audio and vision. Traditionally, both tasks use CNNs; we do not deviate from this approach, and use CNNs for our leaf classifiers as well. Our experimental ecosystem is detailed in Appendix~\ref{app:setup}. We measure robustness and certified accuracy as defined by Cohen \etal~\cite{cohen2019certified}. Through our evaluation, we show that:
\begin{enumerate}
\item Our approach is agnostic to the domain of classification. For both audio and vision tasks, the hierarchical approach shows gains in robustness (through an improvement in certified radius) and accuracy, suggesting that certified accuracy and robustness may no longer need to be at odds.
\item The exact choice of implementation of the root (and in general intermediate classifiers) bears no impact on the correctness of the paradigm. In \S~\ref{casestudy1}, we implement the root classifier as a regular DNN trained to operate on robust features (obtained from a different input modality), and in \S~\ref{casestudy2}, we implement the root classifier as a smoothed classifier~\cite{cohen2019certified}. Both approaches result in a hierarchical classifier with increased robustness and certified accuracy.
\item By adding one additional invariant (location), we observe that there are significant gains in robustness for the vision task (refer \S~\ref{sec:morerobustfeatures}).
\end{enumerate}




\subsection{Road Sign Classification}
\label{casestudy1}

\begin{figure}[h]
  \centering
  \includegraphics[width=\linewidth]{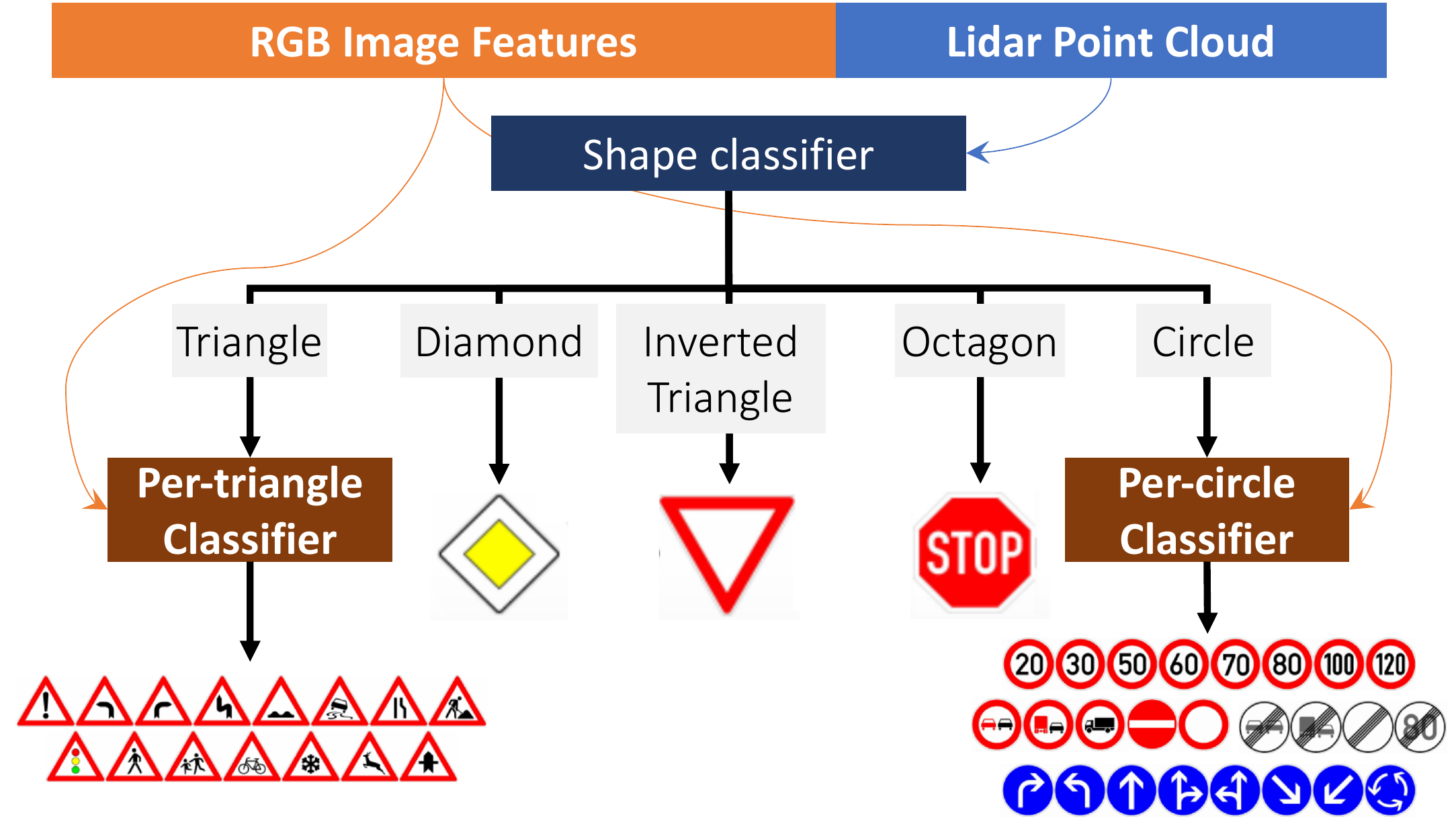}
  \caption{\small The hierarchy over the road signs from the GTSRB dataset.}
  \label{fig:road-sign-setup}
\end{figure}

For road sign classification, we wish to preserve the shape invariance. By doing so, we are able to partition the label space into equivalence classes based on shape (circular, triangular, octagonal, inverse triangular, or rectangular). We first classify road signs first based on their shapes at the root level (using a classifier trained using the ResNet-20 architecture~\cite{DBLP:journals/corr/HeZRS15}). Within each equivalence class \ie per-shape, we perform classification using a smoothed classifier\footnote{Classifier obtained after using the smoothing approach proposed by Cohen \etal~\cite{cohen2019certified}} (which also has the ResNet-20 architecture) to obtain the exact road sign. Note that we set the smoothing noise parameter $\sigma$ set to different values \ie $\sigma=0.25,0.5,1$ to better understand the trade-offs between robustness and certified accuracy. Figure~\ref{fig:road-sign-setup} contains a visual summary of our approach. 

In this particular case study, the hierarchical classification encodes various priors about the classification task which are obtained from domain knowledge about the classification task. To ensure that the invariances (\ie shape) are preserved, we use {\em robust input features} that are obtained from a different sensor. {\em Why shape?} We wished to showcase how domain expertise, and a human understandable invariant improves the robustness certificate. Shape was a natural candidate for such an invariant. 


\subsubsection{Datasets}

\begin{figure}
    \centering
    \includegraphics[width=\linewidth]{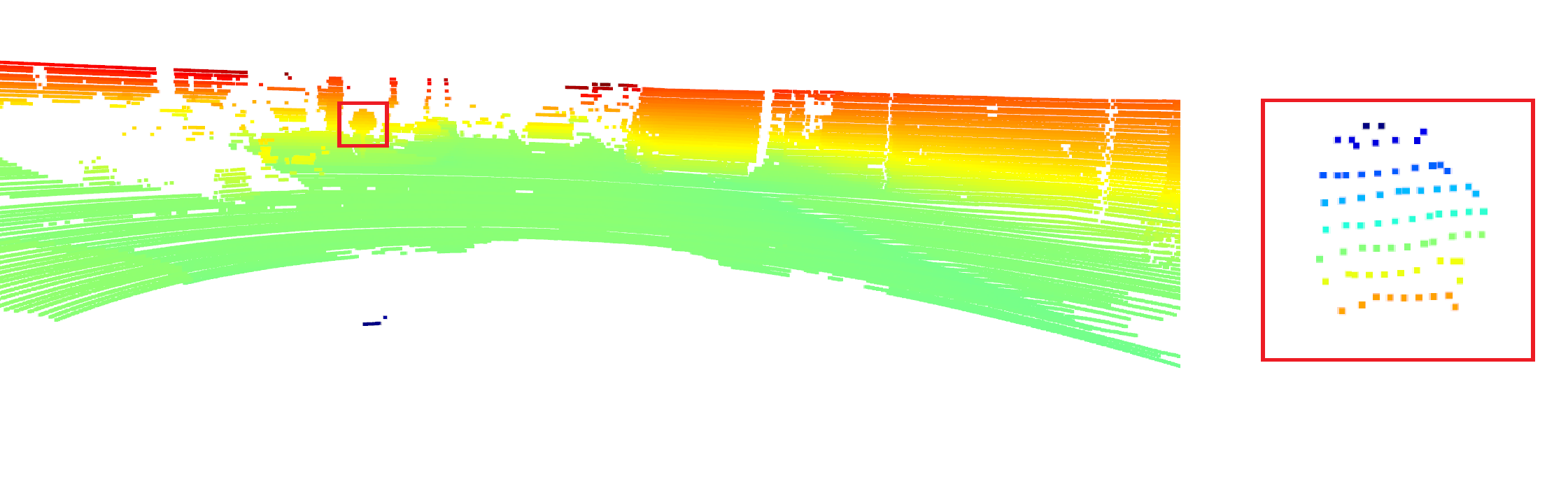}
    \caption{\small Left: A point cloud of a scene in a driving environment. Right: The cropped point cloud of the sign outlined in red in the scene.}
    \label{fig:pc_scene}
\end{figure}

For our experiments, we use two datasets: the first is the German Traffic Sign Recognition Benchmark (GTSRB)~\cite{Stallkamp2012} which contains 51,840 cropped images of German road signs which belong to 43 classes; Figure~\ref{fig:road-sign-setup} shows these classes. The second is the KITTI dataset~\cite{Geiger2013IJRR} which contains time-stamped location measurements, high-resolution images, and \lidar scans over a five-day recording period from an instrumented vehicle on the roads of Karlsruhe, Germany. This totalled 12919 images. We post-processed this dataset to (a) extract only cropped images of the road signs included in the GTSRB dataset, and (b) extract their corresponding \lidar depth maps. To do so, we spent approximately 40 hours manually cropping and annotating every GTSRB road sign found in the KITTI dataset and cropped the corresponding point clouds using the image coordinates. Thus, we obtained 3138 cropped images, their corresponding \lidar point clouds, and their labels (Figure~\ref{fig:pc_scene}). 

We train the root classifier using these \lidar point clouds of road signs to predict the shape of the corresponding road sign. The root has 98.01\% test accuracy. We train each leaf classifier (within an equivalence class) with the road signs belonging to that particular equivalence class. Note that the entire classifier can not be trained using the point clouds as they lack information required to predict the exact road sign label. For example, the shape of 2 \texttt{SpeedLimit} signs can be easily extracted from the corresponding point clouds, but other features required for obtaining the correct label (such as raw pixel intensities) are missing.  

\subsubsection{Attacking the Root Classifier}
\label{improvement}

\begin{figure}[ht]
  \centering
  \includegraphics[width=\linewidth]{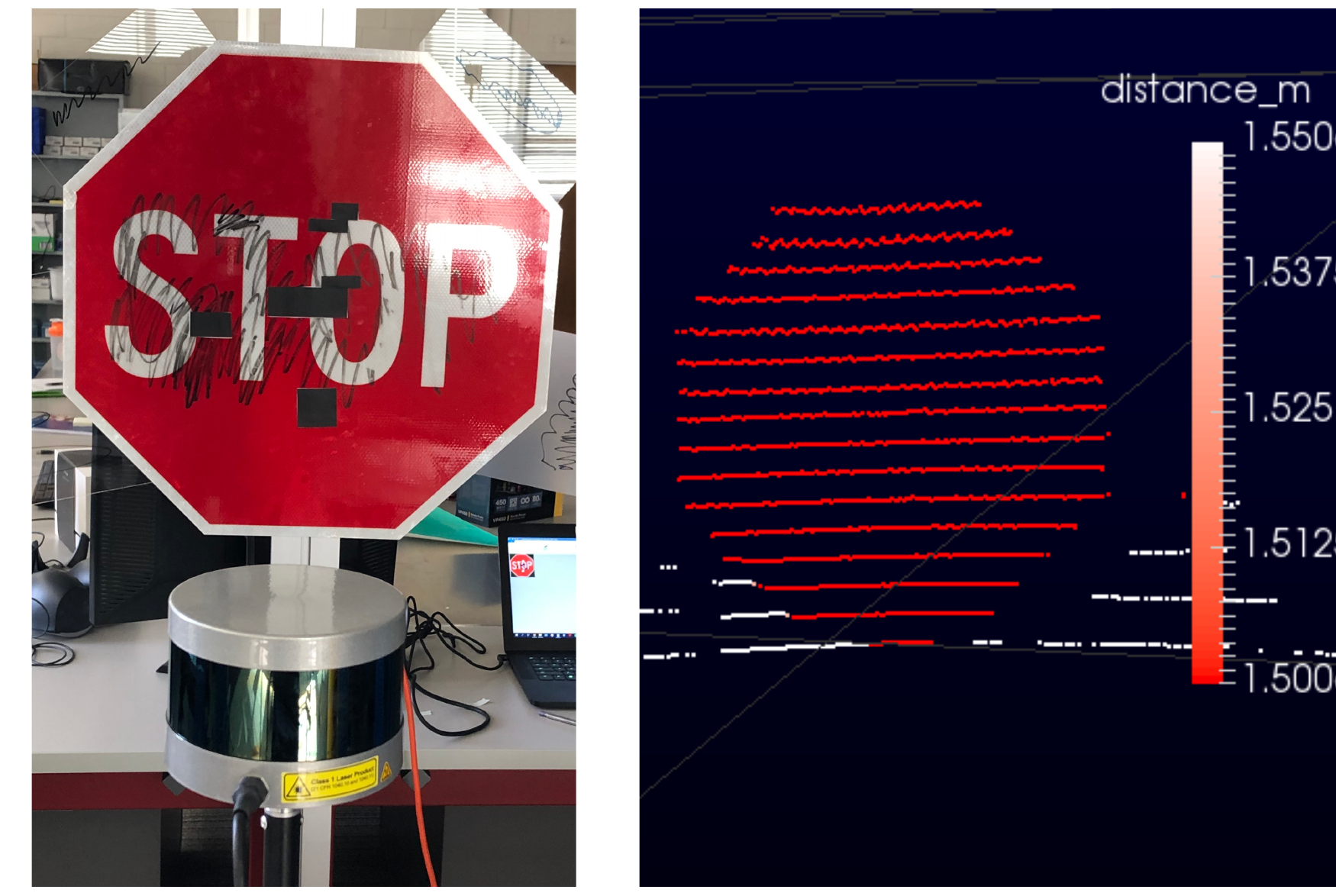}
  \caption{\small The \lidar depth measurements are immune to physical perturbations, including changes in lighting, and any stickers placed on the US road sign.}
  \label{fig:lidar-teaser}
\end{figure}

Human imperceptible perturbations, such as those considered in our threat model, generated for one input modality (such as the pixels obtained from the camera) do not translate over to the other (such as point clouds obtained from the \lidar). Thus, adversaries who can generate $p$-norm bounded perturbations for the road sign (in the pixel space), or deface the road sign with specific stickers do not impact the point clouds corresponding to the road sign. Such attacks (in the pixel space) are the state-of-the-art in literature. We verify our claim by conducting experiments with road signs made of different materials, and a \texttt{Velodyne Puck} \lidar~\cite{velodynepuck}. These experiments suggest that even {\em large} perturbations (such as stickers) made on the road sign do not impact the road sign's point cloud. We display the results from our experiments in Figure~\ref{fig:lidar-teaser}. Observe that our experiments contained far more drastic perturbations than considered in the threat model. Thus, we conclude that the point cloud features are robust to perturbations made to the pixel features. Note that while the \lidar is susceptible to active attacks~\cite{Cao:2019:ASA:3319535.3339815}, such an attack is difficult to mount and are beyond the scope of our threat model (refer \S~\ref{sec:threat}). 

\vspace{1mm}
\noindent{\em Passive attacks on point clouds:} Recent works show $p$-norm bounded adversarial attacks exist for point clouds as well~\cite{point_perturb_generate}. We carry out experiments to understand if these perturbations are human (im)perceptible. We first map the coordinates of each cropped image from the KITTI dataset to its corresponding 3D point cloud, and crop out only the 3D points corresponding to the road sign. By doing so, and with data augmentation methods, we obtain 3642 cropped point clouds of road signs. We then train a PointNet~\cite{pointnet} shape classifier ($\sim$ 98\% test accuracy) using a subset of the point clouds to create a target model. 

We proceed to generate adversarial examples for PointNet using the approach described in ~\cite{point_perturb_generate}. From the results shown in Table~\ref{table:pc_adv}, we see that generating perturbations minimizing the $\ell_2$ distance in the point cloud space is indeed an effective attack. Explicitly attacking the point clouds generated by the \lidar sensor results in (potentially) perceptible perturbations, despite having a small $p$-norm. To the best of our knowledge, there exists no prior work that jointly attacks both sensors\footnote{This approach is commonly referred to as sensor fusion~\cite{gustafsson2010statistical}.} whilst producing small $p$-norm perturbations. We also deploy the clustering attack derived by Xiang \etal~\cite{point_perturb_generate}; the attack generates a cluster of points to resemble a small ball attached to the original object. However, the adversarial clustering attack fails to successfully generate adversarial examples for each of the inputs when attacking majority of the target classes.

Thus, we conclude with the observation that it is possible in theory to generate attacks against the robust feature, these attacks are no longer imperceptible\footnote{Obtained by analyzing the datasheet~\cite{velodynepuck}}; as we state earlier, such attacks are beyond the scope of our threat model.



\begin{table}
\small
\begin{center}
  \begin{tabular}{p{2.3cm}  p{1.2cm}  p{1.2cm}  p{1.2cm} p{1.2cm}}
    \toprule
    \textbf{Output} & \textbf{Circle} & \textbf{Diamond} & \textbf{Triangle} & \textbf{Inverted Triangle}\\
    \midrule
    $||\bar{\delta}||_2$ & 1.032 & 0.636 & 0.386 & 1.059 \\
    $\Delta$ Distance & 3.096 cm & 1.908 cm & 1.158 cm & 3.177 cm \\
    Attack accuracy & 91.33\% & 32.00\% & 98.00\% & 89.33\% \\
    \bottomrule
    \end{tabular}
\caption{\small Results of the attacks described in \S~\ref{improvement}. Legend: (a) $||\bar{\delta}||_2$: Average $\ell_2$ perturbation distance between original and adversarial points, (b) {\bf $\Delta$ Distance:} Coarse upper-bound of changes in depth caused if the perturbation is realized, and (c) \textbf{Attack accuracy:} Percentage of successfully generated perturbation attacks. Observe that some realizations are very perceptible.}
\label{table:pc_adv}
\end{center}
\vspace*{-5mm}
\end{table}



\subsubsection{Retraining vs. Renormalization} 
\label{retrainvsrenormalize}

Under the existence of an accurate and robust root classifier, each leaf classifier only accepts as input road signs belonging to particular shape. As a running example, let us assume that the leaf classifier under discussion classifies {\em circular} road signs. To obtain such a leaf classifier, one could (a) utilize a classifier trained on all labels (henceforth referred to as the baseline classifier) and discard the probability estimates of the labels not of interest (\eg labels of road signs which are not circular), and renormalize the remaining probability estimates; we refer to such an approach as the {\em renormalization} approach, or (b) {\em retrain} a leaf classifier from scratch based on the labels belonging to that particular equivalence class; \eg retrain a classifier for circular road signs in particular. Appendices~\ref{app:circles},~\ref{app:triangles}, and~\ref{app:individual_circles} contains results from both these approaches. We observe that both these approaches increase the robustness certificate; while the renormalization approach can only increase the robustness certificate (by design, we discard the probability estimates of labels we are disinterested in and {\em renormalize} the remaining, ergo widening the gap), the retraining approach can potentially decrease the robustness certificate for some inputs.

Recall that the value of the robustness certificate is directly proportional to the margin $\Phi^{-1}(\underline{p_A})-\Phi^{-1}(\overline{p_B})$, which is dependent on the probability $p_A$ of the top class $c_A$, and probability $p_B$ of the runner-up class $c_B$. Two important observations guide our analysis: (a) note that $c_B$ can either belong to the same equivalence class as $c_A$ (denoted $c_A \approx c_B$), or belong to a different equivalence class (denoted $c_A \napprox c_B$), and (b) the probability estimates for each of the leaf classifiers (which predict only a subset of the labels), before renormalization, are the same as those in the baseline classifier (by construction). On reducing the label space, if $c_A \approx c_B$, renormalization will further widen the margin, and increase the robustness certificate. If  $c_A \napprox c_B$, then there must exist a runner-up candidate $c_{B'} \approx c_A$. Its corresponding runner-up estimate $\overline{p_{B'}} \leq \overline{p_B}$ (or it would have been the runner-up in the first place). Consequently, the margin $\Phi^{-1}(\underline{p_A})-\Phi^{-1}(\overline{p_{B'}}) \geq \Phi^{-1}(\underline{p_A})-\Phi^{-1}(\overline{p_B})$. 


In the retraining scenario, however, we do not have knowledge about the {\em ordering} of the probability estimates in the retrained classifier in comparison to the baseline classifier. It is possible that for a retrained classifier and for a given input, while the correct class' estimate remains $\underline{p_A}$, the new runner-up's estimate $\overline{p_{B'}}$ can be greater, lesser, or equal to the original (baseline) estimate $\overline{p_B}$. Thus, the new robustness certificate can either be lower, greater, or the same as the baseline scenario. This problem is more fundamental; since robustness is a local property relying on the local Lipschitz constant and the structure of the decision spaces, partial or incomplete knowledge of any of these can result in spurious selection of the runner-up label. An added benefit of the renormalization approach is that it is computationally efficient; one needs to train one baseline classifier (independent of how the label space is partitioned) as opposed to several leaf classifiers (depending on the nature of the partition). 


\vspace{1mm}
\noindent{\bf Results:} In Tables~\ref{table:case_study_11} and~\ref{table:case_study_12}, we present the results of our evaluation. We report the mean and standard deviation of the robustness certificate obtained for different values of $\sigma$. For each equivalence class, we use 3000 inputs belonging to that particular equivalence class (\eg we use 3000 images of circular road signs to validate the improvements in Table~\ref{table:case_study_11}). We also report the certified accuracy for varying values of $\sigma$ (refer Appendix D in ~\cite{cohen2019certified} for the detailed definition and formulation). In each table, the columns pertaining to the baseline contain the certificates (and corresponding certified accuracy) for only the labels belonging to a particular equivalence class (circles in Table~\ref{table:case_study_11} and triangles in Table~\ref{table:case_study_12}). We can see that by simply rearchitecting the classifier to enforce invariances (in this case, obtained using auxiliary inputs), we are able to improve the certified radius and certified accuracy\footnote{For brevity, we only report results for individual equivalence classes; the overall certified accuracy of both the baseline and our hierarchical approach is comparable.}. As stated earlier, more fine-grained results are presented in Appendices~\ref{app:triangles},~\ref{app:circles}, and~\ref{app:individual_circles}.

\begin{table}
\small
\begin{center}
  \begin{tabular}{p{1.25cm}|  p{1.4cm}|  p{1.5cm}|  p{1.25cm}| p{1.5cm}}
    \toprule
    \textbf{}       & {\footnotesize Baseline CR} & {\footnotesize Hierarchy CR}   & {\footnotesize Baseline CA} & {\footnotesize Hierarchy CA}\\
    \midrule
    $\sigma=0.25$   & 1.083 $\pm$ 0.579 & 1.090 $\pm$ 0.575  & 77.74\%    &  79.25\%\\
    $\sigma=0.5$    & 1.620 $\pm$ 0.986 & 1.642 $\pm$ 1.009  & 56.54\%    &  58.39\%\\
    $\sigma=1.0$    & 2.327 $\pm$ 1.454 & 2.400 $\pm$ 1.560    & 31.69\%    &  35.47\%\\
    \bottomrule
    \end{tabular}
\caption{\small For inputs that belong to the circular equivalence class, the hierarchical approach preserving invariances outperforms the baseline model.}
\label{table:case_study_11}
\end{center}
\vspace*{-6mm}
\end{table}

\begin{table}
\small
\begin{center}
  \begin{tabular}{p{1.25cm} | p{1.4cm} | p{1.5cm} | p{1.25cm} | p{1.5cm}}
    \toprule
    \textbf{}       & {\footnotesize Baseline CR} & {\footnotesize Hierarchy CR}   & {\footnotesize Baseline CA} & {\footnotesize Hierarchy CA}\\
    \midrule
    $\sigma=0.25$   & \textcolor{black}{1.107 $\pm$ 0.551}  & 1.127 $\pm$ 0.547   & 71.60\% & 76.45\% \\
    $\sigma=0.5$    & \textcolor{black}{1.641 $\pm$ 0.919}  & 1.659 $\pm$ 0.899   & 50.67\% & 58.94\% \\
    $\sigma=1.0$    & \textcolor{black}{2.223 $\pm$ 1.259}  & 2.233 $\pm$ 1.172   & 20.41\% & 32.78\% \\
    \bottomrule
    \end{tabular}
\caption{\small For inputs that belong to the triangular equivalence class, the hierarchical approach preserving invariances outperforms the baseline model.}
\label{table:case_study_12}
\end{center}
\vspace*{-6mm}
\end{table}

\subsubsection{Multiple Invariants}
\label{sec:morerobustfeatures}

\begin{figure}[ht]
  \centering
  \includegraphics[width=0.55\linewidth]{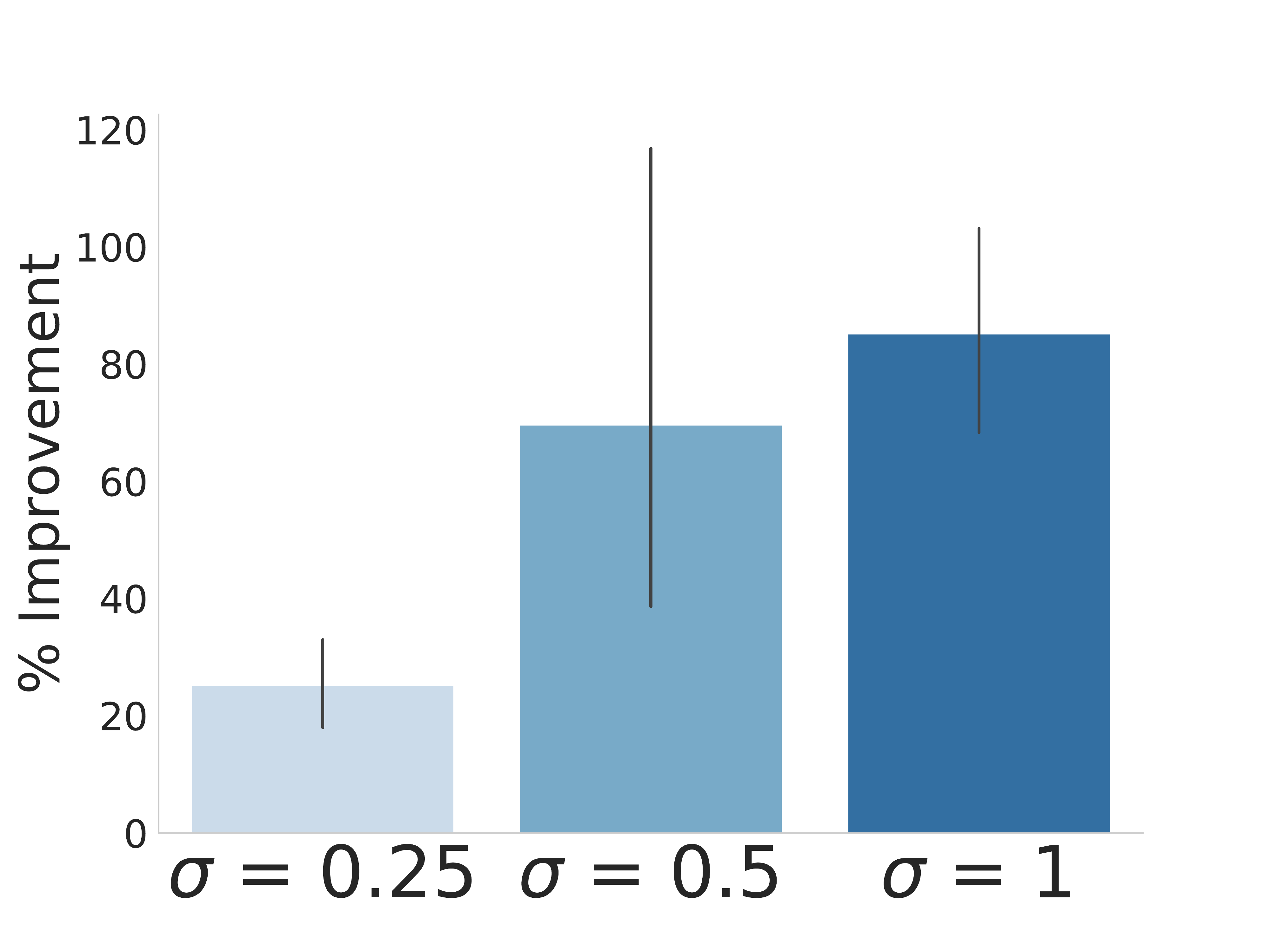}
  \caption{\small The percentage improvement of the robustness certificate for \texttt{SpeedLimit} signs utilizing 2 invariances - shape and location. These results are a significant improvement on those of Table~\ref{table:case_study_11}. Additional invariances do not impact the accuracy in this case study.}
  \label{fig:speedlimits}
\end{figure}

As discussed earlier, the hierarchical classification paradigm can support enforcing more than a single invariant. In this case study, for example, one can further partition the label space using location information. For example, a highway can not have stop signs, or an intersection will not have a speed limit sign etc. To validate this hypothesis, we performed a small scale proof-of-concept experiment to further constrain the space of labels that is obtained by splitting on the shape invariance (\ie see if we can obtain a subset of the set of, say, circular labels). Using the location information from the KITTI dataset, and local map data from OpenStreetMap~\cite{OSM}. For particular locations, we can further constrain the space of circular road signs to just \texttt{SpeedLimit} signs. From Figure~\ref{fig:speedlimits}, we observe that increasing the number of invariances (to 2 - shape {\em and location}) increases the robustness certificate further without impacting accuracy. 

\subsection{Speaker Classification}
\label{casestudy2}


Thus far, our results suggest that hierarchical classification and invariances improve robustness in the {\em vision domain}. We carry out another case study in the audio domain to verify that our approach is general. To this end, we use speaker identification as the next case study. Conventional speaker identification systems utilize a siamese architecture~\cite{chen2011extracting}; speech samples are first converted to discrete features, and these are then converted to an embedding (refer footnote 9). The distance between an unknown embedding (which is to be labeled), and embeddings in a known support set are computed. The label of the known embedding in the support set which is at the least distance from the unknown embedding is returned as the label for the unknown sample. Observe that such an architecture is not conducive for computing certificates using randomized smoothing; the output returned by the network is a label, and there is no associated probability estimate for this prediction. Thus, we implement speaker identification using a conventional CNN comprising of 3 convolutional layers followed by a fully connected layer and a softmax layer. For the remainder of this section, we assume that the inputs are discrete features obtained after processing the voice samples.

As before, we classify inputs based on the gender of the speaker. Since gender is binary, we utilize smoothed DNN (with the same architecture as above) for binary classification. Unlike the previous case study where the root is robust to perturbation as its inputs are robust, the robustness guarantees for this case study come from the design of the classifier (through randomized smoothing~\cite{cohen2019certified}). While we could utilize another modality to obtain speaker gender, we use a robust classifier to highlight the flexibility of our approach. As before, the leaves are randomly smoothed classifiers. 

\subsubsection{Dataset}

For our experiments, we use the LibriSpeech ASR corpus~\cite{panayotov2015librispeech} which comprises of over 1000 hours of recordings of both male and female speakers. In particular, we use 105894 audio samples corresponding to 1172 speakers (606 male speakers and 566 female speakers). These audio samples were first downsampled (to 4 KHz) and then cut to lengths of 3 seconds each. 

\noindent{\bf Results:} As in \S~\ref{retrainvsrenormalize}, we renormalized the outputs of the baseline classifier (trained on inputs from both male and female speakers) to obtain the estimates required for the robustness certificate. Unlike the vision scenario, increasing the standard deviation of the noise added while training the smoothed classifier resulted in poorly performing classifiers (with very low test accuracy and certified accuracy). Thus, compared to the case study in \S~\ref{casestudy1}, the values of $\sigma$ are considerably lower. For each equivalence class, we evaluate our approach with 3000 samples belonging to that class. We observe a considerable increase in the robustness certificate - 1.92$\times$ for male speakers (refer Table~\ref{table:case_study_22}), and 2.07$\times$ for female speakers (refer Table~\ref{table:case_study_21}). As before, we observe an increase in the certified accuracy as well. This suggests that the choice of the invariant is very important, and can enable significant gains in robustness radius without degrading classification accuracy. In the sections that follow, we discuss this point in greater detail.

\begin{table}
\small
\begin{center}
  \begin{tabular}{p{1.4cm}|  p{1.25cm}|  p{1.4cm}|  p{1.5cm}| p{1.25cm}}
    \toprule
    \textbf{}       & \text{\footnotesize Baseline CR} & \text{\footnotesize Hierarchy CR}   & \text{\footnotesize Baseline CA} & \text{\footnotesize Hierarchy CA}\\
    \midrule
    $\sigma=0.025$   & 0.1233 $\pm$ 0.0424 & 0.2554 $\pm$ 0.0429  & 88.80\%    &  89.62\%\\
    $\sigma=0.05$    & 0.1855 $\pm$ 0.0907 & 0.3843 $\pm$ 0.0915  & 65.70\%    &  66.98\%\\
    $\sigma=0.1$     & 0.2612 $\pm$ 0.1579 & 0.5688 $\pm$ 0.1576  & 24.79\%    &  24.96\%\\
    \bottomrule
    \end{tabular}
\caption{\small For inputs that belong to the female gender equivalence class, the hierarchical approach preserving invariances outperforms the baseline model.}
\label{table:case_study_21}
\end{center}
\vspace*{-6mm}
\end{table}

\begin{table}
\small
\begin{center}
  \begin{tabular}{p{1.4cm}|  p{1.25cm}|  p{1.4cm}|  p{1.5cm}| p{1.25cm}}
    \toprule
    \textbf{}       & \text{\footnotesize Baseline CR} & \text{\footnotesize Hierarchy CR}   & \text{\footnotesize Baseline CA} & \text{\footnotesize Hierarchy CA}\\
    \midrule
    $\sigma=0.025$   & 0.1351 $\pm$ 0.0381 & 0.2590 $\pm$ 0.0379  & 92.67\%    &  \textcolor{black}{92.65\%}\\
    $\sigma=0.05$    & 0.2085 $\pm$ 0.0911 & 0.4014 $\pm$ 0.0908  & 76.58\%    &  78.85\%\\
    $\sigma=0.1$     & 0.2883 $\pm$ 0.1563 & 0.5562 $\pm$ 0.1592  & 36.92\%    &  \textcolor{black}{36.40\%}\\
    \bottomrule
    \end{tabular}
\caption{\small For inputs that belong to the male gender equivalence class, the hierarchical approach preserving invariances outperforms the baseline model.}
\label{table:case_study_22}
\end{center}
\vspace*{-6mm}
\end{table}

%% file: Contents/7_discussion.tex
\section{Discussion}
\label{sec:discussion}


\subsection{Do invariances always exist?} 

Recent work \cite{new_madry_2019} suggests that robust features are a by-product of adversarial training. However, extracting such robust features is computationally expensive. In our approach, we use a combination of domain knowledge and observing intermediary representations of input features to extract invariances. But it is unclear if these approaches work for {\em all} classification tasks. Discarding the approaches specified in our work, it is an open problem to determine an algorithm to {\em efficiently} extract invariances/stable properties pertaining to any learning task.

It is also unclear if there is explicit benefit from having invariances that can be defined in a manner that is interpretable to humans. Interpretability is subjective and enforcing such subjectivity might be detrimental to learning (and generalization); such subjectivity might also induce issues pertaining to fairness. 

\subsection{How do we handle multiple invariances?}

In our evaluation thus far, we have discussed how a single invariance can enable robustness without a detriment to accuracy. In the presence of multiple such invariances, it is unclear how they might interact. To begin with, it is unclear if all invariances contribute equally towards robustness, or if some invariances are preferred to others. The same question exists in terms of the interaction between the invariances and accuracy. It is also unclear how one might combine multiple invariances to create a hiearachy with increased robustness in a meaningful manner \ie the ordering of invariances is an open problem. This stems from the fact that it is a challenging proposition understanding the information gain from these invariances.

\subsection{Can better classification architectures improve robustness further?} 

We empirically validate the benefits of using a tree-like hierarchy to improve robustness. However, it is unclear if our approach of constructing the hierarchy is the {\em only} way. Indeed, there are other approaches that can be used to construct classification architectures, such as those discussed in works in the neural architectural search domain~\cite{zoph2016neural}. An interplay of these ideas (searching for architectures that explicitly improve robustness by enforcing invariances) makes for interesting future research. 

\subsection{Do invariances always help?}

Recent research~\cite{DBLP:journals/corr/abs-1903-10484, jacobsen2018excessive} suggests that by default, DNNs are invariant to a wide range of task-related changes. Thus, while trying to enforce invariances which we believe may improve robustness, we could potentially introduce a new region of vulnerabilities. Understanding if specific types of invariances are beneficial or harmful, and where to draw the line is avenue for future work.

%% file: Contents/8_related_work.tex
\section{Related Work}
\label{sec:related_work}

Researchers have extensively studied the robustness of Machine Learning models through exploring new attack strategies and various defense mechanisms. These efforts are very well documented in literature~\cite{DBLP:journals/corr/abs-1902-06705}. In this section, we only discuss work related to the different components of our classification pipeline. 


\paragraph{Hierarchical Classification}

Recent research casts image classification as a visual recognition task ~\cite{HD-CNN,category_structure,treepriors_transferlearning}. The common observation is that these recognition tasks introduce a hierarchy; enforcing a hierarchical structure further improves the accuracy. Similar to our approach, Yan \etal ~\cite{HD-CNN} propose a HD-CNN that classifies input images into coarse categories which then pass corresponding leaf classifiers for fine-grained labeling. They perform spectral clustering on the confusion matrix of a baseline classifier to identify the clusters of categories. This approach is optimized for natural accuracy and uses the image data at all levels of hierarchy. In contrast, we employ robust features from different modalities to construct more robust classifiers.  

Srivastava \etal ~\cite{treepriors_transferlearning} show that leveraging the hierarchical structure can be very useful when there is limited access to inputs belonging to certain classes. They propose an iterative method which uses training data to optimize the model parameters and validation data to select the best tree starting from an initial pre-specified tree. This approach further motivates our tree-based hierarchy; in several settings, such as autonomous driving systems, a hierarchy is readily available (as displayed by our experiments with shape and location).

\paragraph{Imperceptible Adversarial Examples in the Wild}

Extensive research is aimed at generating digital adversarial examples, and defenses corresponding to $p$-norm bounded perturbations to the original inputs ~\cite{goodfellow2014explaining,papernotattack,kurakin2016adversarial,madry-iclr2018}. However, these studies fail to provide robustness guarantees for the attacks realizable in the physical world due to a variety of factors including view-point shifts, camera noise, domain adaptation, and other affine transformations.

The first results in this space were presented by Kurakin \etal~\cite{kurakin2016adversarial}. The authors generate adversarial examples for an image, print them, and verify if the prints are adversarial or not.  Sharif \etal developed a physical attack approach~\cite{sharif2016accessorize, sharif2017adversarial} on face recognition systems using a printed pair of eyeglasses. Recent work with highway traffic signs demonstrates that both state-of-the-art machine learning classifiers, as well as detectors, are susceptible to real-world physical perturbations~\cite{roadsigns17,object-detector-attacks}. Athalye \etal~\cite{AthalyeS17} provide an algorithm to generate 3D adversarial examples (with small $p$-norm), relying on various transformations (for different points-of-view).

\paragraph{\lidar Attacks}

Similar to our approach, Liu \etal~\cite{pointcloud} adapt the attacks and defense schemes from the 2D regime to 3D point cloud inputs. They have shown that even simpler defenses such as outlier removal, and removing salient points are effective in safeguarding point clouds. This observation further motivates our selection of point clouds as auxiliary inputs in the case study. However, Liu \etal ~\cite{pointcloud} do not physically realize the generated perturbations. Other approaches consider active adversarial attacks against the \lidar modalities~\cite{active_lidar}, which can be expensive to launch. In this paper, we focus on passive attacks (on sensors) through object perturbations.

Xiang \etal ~\cite{point_perturb_generate} propose several algorithms to add adversarial perturbations to point clouds through  generating new points or perturbing existing points. An attacker can generate an adversarial point cloud, but manifesting this point cloud in the physical world is a different story. There are several constraints need to be accounted for, such as the \lidar's vertical and horizontal resolution and the scene's 3D layout. . Still, an attacker would need to attack more than one modality to cause a misclassification. 

\paragraph{Robust Features}

Ilyas \etal ~\cite{new_madry_2019} and Tsipras \etal ~\cite{tsipras2018there} distinguish robust features from non-robust features to explain the trade-off between adversarial robustness and natural accuracy. While the authors show an improved trade-off between standard accuracy and robust accuracy, it is achieved at the computational cost of generating a large, robust dataset through adversarial training ~\cite{new_madry_2019}. We circumvent this computational overhead by adopting invariants (and consequently robust features) imposed by the constraints in the physical world. 


%% file: Contents/9_conclusion.tex
\section{Conclusion}
\label{sec:conclusion}

In this paper, we discuss how robust features realized through invariances (obtained through domain knowledge, or provided by real world constraints), when imposed on a classification task can be leveraged to significantly improve adversarial robustness without impacting accuracy. Better still, this is achieved at minimal computational overhead. Through a new hierarchical classification approach, we validate our proposal on real-world classification tasks -- road sign classification and speaker identification. We also show how some invariances can be used to safeguard the aforementioned classification task from physically realizable adversarial examples (in the case of road sign classification). 




%% file: Contents/appendix.tex
\section{Tables}

\subsection{Experimental Setup}
\label{app:setup}

All experiments were conducted on two servers. The first server has 264 GB memory, 8 NVIDIA’s GeForce RTX 2080 GPUs, and 48 CPU cores. The second has 128GB memory, 2 NVIDIA's TITAN xp GPUs $\And$ 1 NVIDIA's Quadro P6000 GPU, and 40 CPU cores.

\subsection{Settings for Experiments in \S~\ref{sec:insights}}
\label{app:insights}

Projected Gradient Descent (PGD) is a white box attack bounded by the perturbation size, $\epsilon$. The results derived in \S~\ref{sec:insights} were evaluated on PGD with 20 iterations and 40 iterations (PGD-20, PGD-40) on the $\infty$-norm~\cite{madry-iclr2018}.  Since adversarial training is notoriously slow (as it attempts to solve a nested optimization), we utilize YOPO~\cite{yopo}, a technique which accelerates adversarial training by up to 4-5 $\times$. This reduces the amount of times the network is propagated while staying competitive with the results of typical adversarial training using PGD. We perform our adversarial training experiments using YOPO-5-3 so that only 5 full forward and backward-propagations are made instead of the 20 with PGD-20. The same is done with YOPO-5-10 for approximating PGD-40.

\begin{table}[ht]
\small
\begin{center}
  \begin{tabular}{p{1.4cm}  p{1.4cm}  p{0.4cm}  p{0.4cm} p{0.4cm} p{0.5cm} p{0.5cm} p{0.4cm} p{0.3cm}}
    \toprule
    \text{\footnotesize Model} & \text{\footnotesize Architecture} & \text{n} & \textbf{$\sigma$} & \textbf{$\varepsilon$} & \textbf{$\kappa$} & \textbf{$\eta$} & \text{b}\\
    \midrule
    {\footnotesize YOPO-5-3} &  {\footnotesize ResNet-34}  & {\footnotesize 200} & {\footnotesize -} & {\footnotesize $\frac{8}{255}$} & {\footnotesize $5\mathrm{e}{-2}$} & {\footnotesize $2\mathrm{e}{-3}$} & {\footnotesize 128} \\
    {\footnotesize YOPO-5-10} &  {\footnotesize Small CNN}  & {\footnotesize 40} & {\footnotesize -} & {\footnotesize $0.47$} & {\footnotesize $5\mathrm{e}{-2}$} & {\footnotesize $2\mathrm{e}{-3}$} & {\footnotesize 128} \\
    \midrule
    {\footnotesize Smoothing} & {\footnotesize ResNet-110} & {\footnotesize 350} & {\footnotesize 0.25} & {\footnotesize -} & {\footnotesize $1\mathrm{e}{-2}$} & {\footnotesize $1\mathrm{e}{-2}$} & {\footnotesize 256} \\
    {\footnotesize Smoothing} & {\footnotesize ResNet-110} & {\footnotesize 350} & {\footnotesize 0.50} & {\footnotesize -} & {\footnotesize $1\mathrm{e}{-2}$} & {\footnotesize $1\mathrm{e}{-2}$} & {\footnotesize 256} \\
    {\footnotesize Smoothing} & {\footnotesize ResNet-110} & {\footnotesize 350} & {\footnotesize 1.00} & {\footnotesize -} & {\footnotesize $1\mathrm{e}{-2}$} & {\footnotesize $1\mathrm{e}{-2}$} & {\footnotesize 256} \\
    \bottomrule
    \end{tabular}
\caption{\small The training parameters used for our YOPO and randomized smoothing models. YOPO-5-3 (with step size $\frac{2}{255}$) and smoothing models were used for obtaining results on CIFAR-10 dataset. YOPO-5-10 (with step size $0.01$) models were used for obtaining results on MNIST dataset. All of the parameters except for the number of labels were kept consistent when training our baseline, leaf, and root classifiers for our experiments in \S~\ref{sec:insights}. (a) n: number of samples, (b) $\sigma$: standard deviation of noise added for smoothing, (c) $\varepsilon$: maximum permissible noise added while adversarially training, (d) $\kappa$: weight decay, (e) $\eta$: learning rate, and (f) b: batch size.}
\label{table:parameters}
\end{center}
\end{table}

\begin{table}[ht]
\small
\begin{center}
  \begin{tabular}{p{1cm}  p{1.5cm}  p{1.5cm}  p{1cm}}
    \toprule
    \text{\footnotesize Model} & \text{\footnotesize Natural (\%)} & \text{\footnotesize Adv (\%)} & \text{\footnotesize \# Classes}\\
    \midrule
    {\footnotesize Root} & {\footnotesize 98.51\%} & {\footnotesize 93.68\%} & {\footnotesize 2}\\
    {\footnotesize 4Class} & {\footnotesize 98.37\%} & {\footnotesize 86.95\%} & {\footnotesize 4}\\
    {\footnotesize 6Class} & {\footnotesize 98.52\%} & {\footnotesize 85.81\%} & {\footnotesize 6}\\
    {\footnotesize Baseline} & {\footnotesize 97.86\%} & {\footnotesize 82.35\%} & {\footnotesize 10}\\
    \bottomrule
    \end{tabular}
\caption{\small The results on the test set for each YOPO-5-10 model used in our evaluation for MNIST. Using the parameters from Table~\ref{table:parameters}, we train a root classifier, 2 leaf classifiers, and 1 baseline classifier trained on the entire label space. The legends are (a) {\bf Natural (\%)}: natural accuracy, (b) {\bf Adv (\%)}: adversarial accuracy with PGD-40 attack where $\sigma=0.01, \epsilon=0.47$}
\label{table:mnist_acc}
\end{center}
\end{table}

In Table~\ref{table:cifar_acc} and Table~\ref{table:mnist_acc}, we report statistics for the experiments related to MNIST and CIFAR-10 in \S~\ref{sec:insights}. Specifically, we report the natural  and adversarial accuracy of different constitutent classifiers are reported. Trained with the parameters from Table~\ref{table:parameters}, these classifiers were used in our end to end evaluation of our hierarchical classification scheme.

\begin{table}[ht]
\small
\begin{center}
  \begin{tabular}{p{1.5cm}  p{1.5cm}  p{1.5cm}  p{1cm}}
    \toprule
    \text{\footnotesize Model} & \text{\footnotesize Natural (\%)} & \text{\footnotesize Adv (\%)} & \text{\footnotesize \# Classes}\\
    \midrule
    {\footnotesize Root} & {\footnotesize 96.57\%} & {\footnotesize 84.25\%} & {\footnotesize 2}\\
    {\footnotesize Non-living} & {\footnotesize 93.48\%} & {\footnotesize 69.04\%} & {\footnotesize 4}\\
    {\footnotesize Living} & {\footnotesize 80.72\%} & {\footnotesize 40.78\%} & {\footnotesize 6}\\
    {\footnotesize Baseline} & {\footnotesize 83.68\%} & {\footnotesize 38.77\%} & {\footnotesize 10}\\
    \bottomrule
    \end{tabular}
\caption{\small The results on the test set for each YOPO-5-3 model used in our evaluation for CIFAR-10. Using the parameters from Table~\ref{table:parameters}, we train a root classifier, 2 leaf classifiers, and 1 baseline classifier trained on the entire label space. The legends are (a) {\bf Natural (\%)}: natural accuracy, (b) {\bf Adv (\%)}: adversarial accuracy with PGD-20 attack where $\sigma=\frac{2}{255}, \epsilon=\frac{8}{255}$}
\label{table:cifar_acc}
\end{center}
\end{table}

\subsection{Robustness and Accuracy no Longer at Odds}
\label{tsipras}

We consider the setting of Tsipras et al.~\cite{tsipras2018there,new_madry_2019} of a binary classification task. In this setting, robustness and accuracy are at odds~\cite{tsipras2018there}. Here, we show, over the same setting, that imposing an invariant on the attacker improves the defender's accuracy-robustness trade-off.

\input{ccs-2019/new_madry_stuff}

\subsection{Data Driven Invariances}
\label{app:data}

\begin{figure}[ht]
  \centering
  \includegraphics[width=\linewidth]{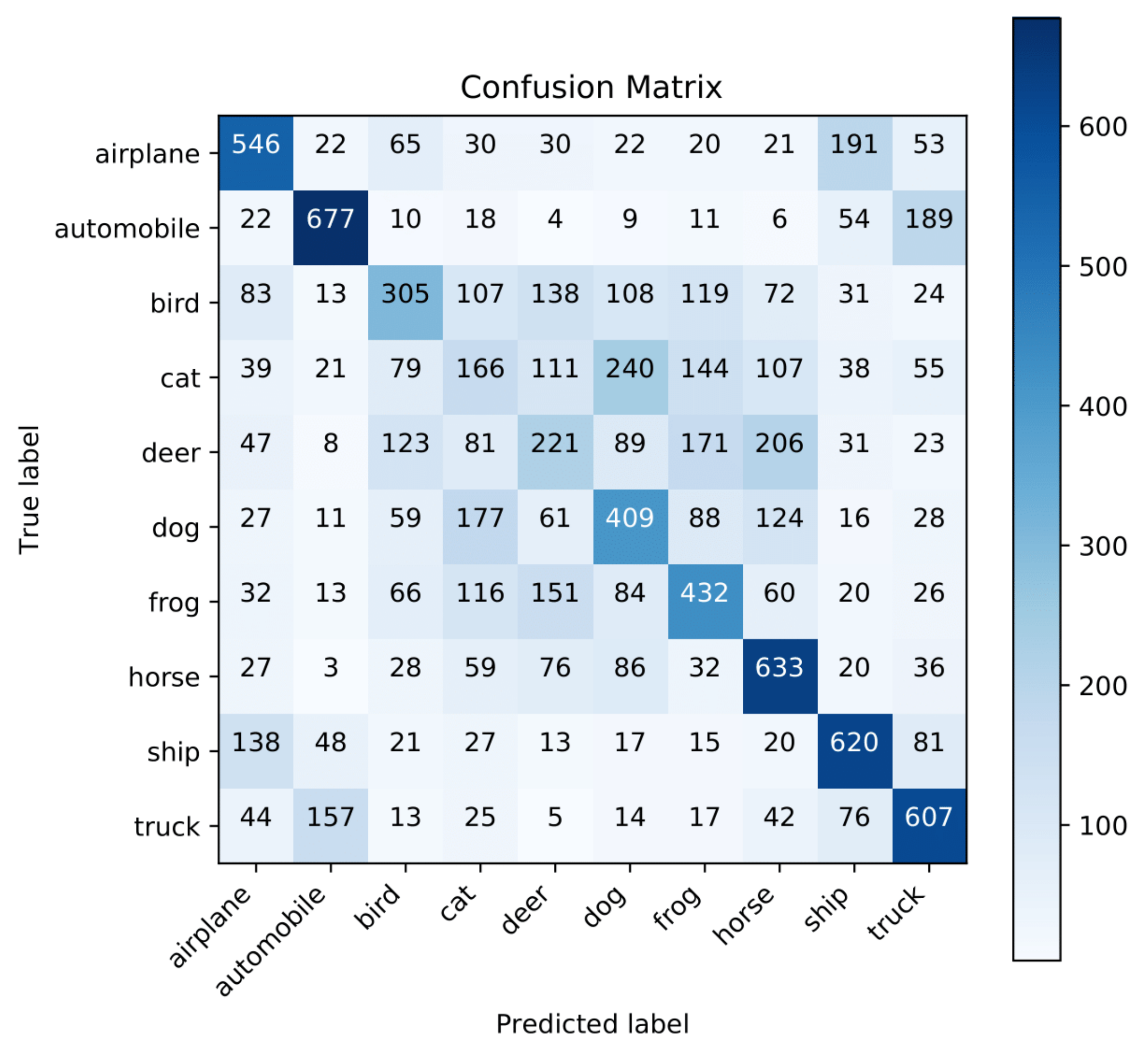}
  \caption{\small Confusion matrix obtained by inspecting the predictions of an adversarially trained ($\epsilon=\frac{8}{255})$ CNN.}
  \label{fig:confusionmatrix1}
\end{figure}

By observing the confusion matrix, one can see that for the CIFAR-10 dataset, Non-living objects (such as planes, automobiles, ships, and trucks) are more likely to be misclassified among each other, in comparison to being misclassifed as Living objects (such as animals). This motivates our split as described in \S~\ref{sec:obtaining}.

\subsection{CIFAR-10 Certification}
\label{app:cifar}

Instead of retraining the leaf classifiers with adversarial training, we retrain the leaf classifiers on the CIFAR-10 dataset with randomized smoothing. We still utilize the root classifier as described in \S~\ref{sec:method}. From Table~\ref{table:vehicles_ca} and Table~\ref{table:animals_ca}, we can see that the hierarchical approach increases the robustness certificate and certified accuracy, in comparison to the flat baseline. 
\begin{table}[ht]
\small
\begin{center}
  \begin{tabular}{p{1.25cm} | p{1.4cm} | p{1.5cm} | p{1.25cm} | p{1.5cm}}
    \toprule
    \textbf{}       & {\footnotesize Baseline CR} & {\footnotesize Hierarchy CR}   & {\footnotesize Baseline CA } & {\footnotesize Hierarchy CA}\\
    \midrule
    {\footnotesize $\sigma=0.25$}   & 0.6749 $\pm$ 0.2798 & 0.7204 $\pm$ 0.2692 & 88.15\% & 90.70\% \\
    {\footnotesize $\sigma=0.5$}    & 0.9270 $\pm$ 0.5349 & 1.0943 $\pm$ 0.5686 & 79.58\% & 85.85\% \\
    {\footnotesize $\sigma=1.0$}    & 1.2270 $\pm$ 0.7789 & 1.5013 $\pm$ 0.9828 & 65.23\% & 74.74\% \\
    \bottomrule
    \end{tabular}
\caption{\small For inputs that belong to the Non-Living objects equivalence class, the hierarchical approach preserving invariances outperforms the baseline model.}
\label{table:vehicles_ca}
\end{center}
\vspace*{-4mm}
\end{table}

\begin{table}[ht]
\small
\begin{center}
  \begin{tabular}{p{1.25cm} | p{1.4cm} | p{1.5cm} | p{1.25cm} | p{1.5cm}}
    \toprule
    \textbf{}       & \text{\footnotesize Baseline CR} & \text{\footnotesize Hierarchy CR}   & \text{\footnotesize Baseline CA} & \text{\footnotesize Hierarchy CA}\\
    \midrule
    $\sigma=0.25$   & 0.5263 $\pm$ 0.2987 & 0.5671 $\pm$ 0.2966 & 75.93\% & 80.80\% \\
    $\sigma=0.5$    & 0.7219 $\pm$ 0.5037 & 0.7777 $\pm$ 0.5140 & 61.82\% & 67.05\% \\
    $\sigma=1.0$    & 0.9319 $\pm$ 0.7486 & 1.0750 $\pm$ 0.8278 & 42.68\% & 53.80\% \\
    \bottomrule
    \end{tabular}
\caption{\small For inputs that belong to the Living objects equivalence class, the hierarchical approach preserving invariances outperforms the baseline model.}
\label{table:animals_ca}
\end{center}
\vspace*{-4mm}
\end{table}

\subsection{Retraining vs. Renormalization for Triangular Road Signs}
\label{app:triangles}

\begin{figure}[ht]
  \centering
  \includegraphics[width=0.5\linewidth]{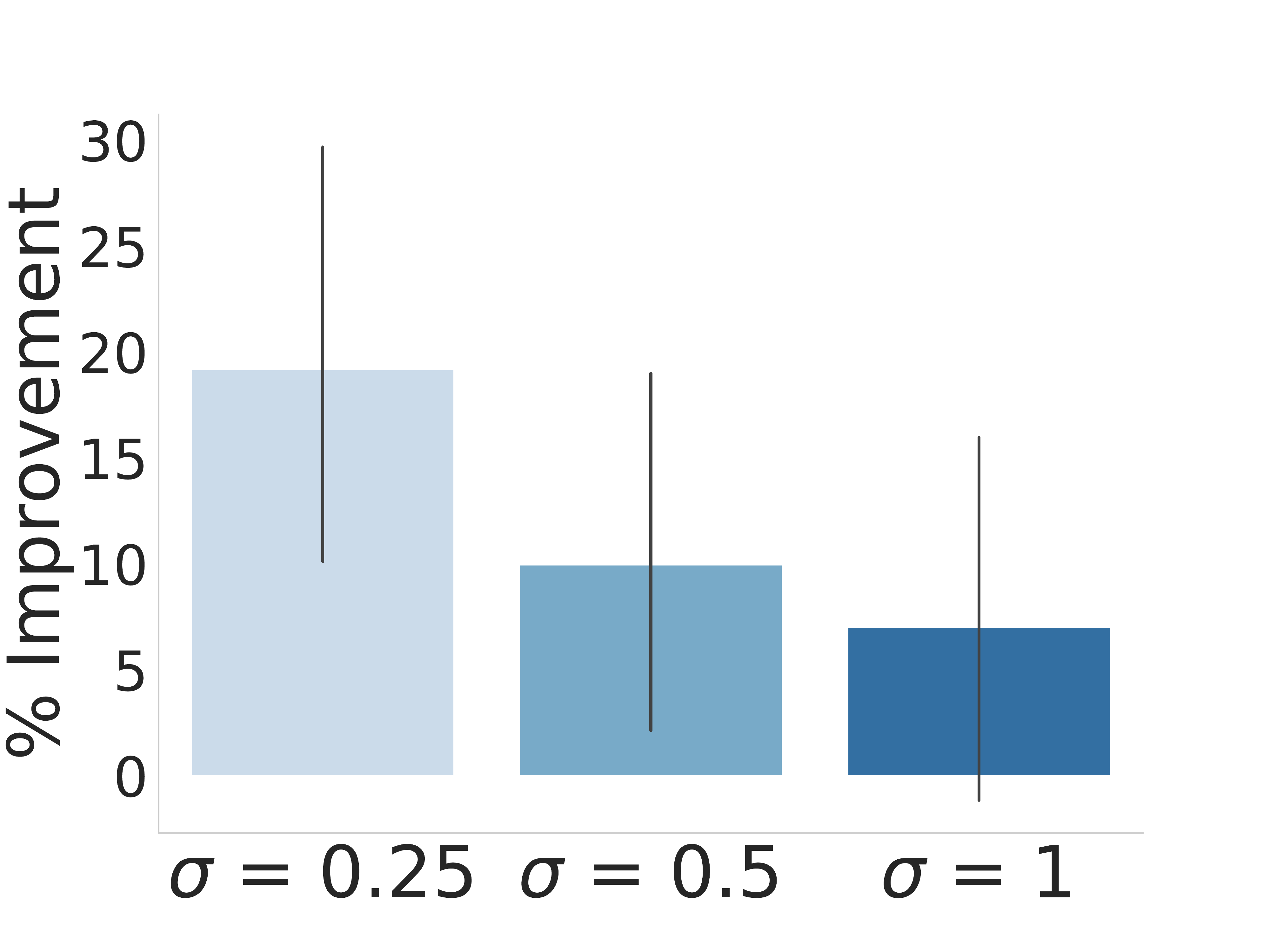}
  \caption{\small Improvements in robustness when the triangle leaf classifier is {\em retrained}.}
  \label{fig:triangle_retrain}
\end{figure}

\begin{figure}[ht]
  \centering
  \includegraphics[width=0.5\linewidth]{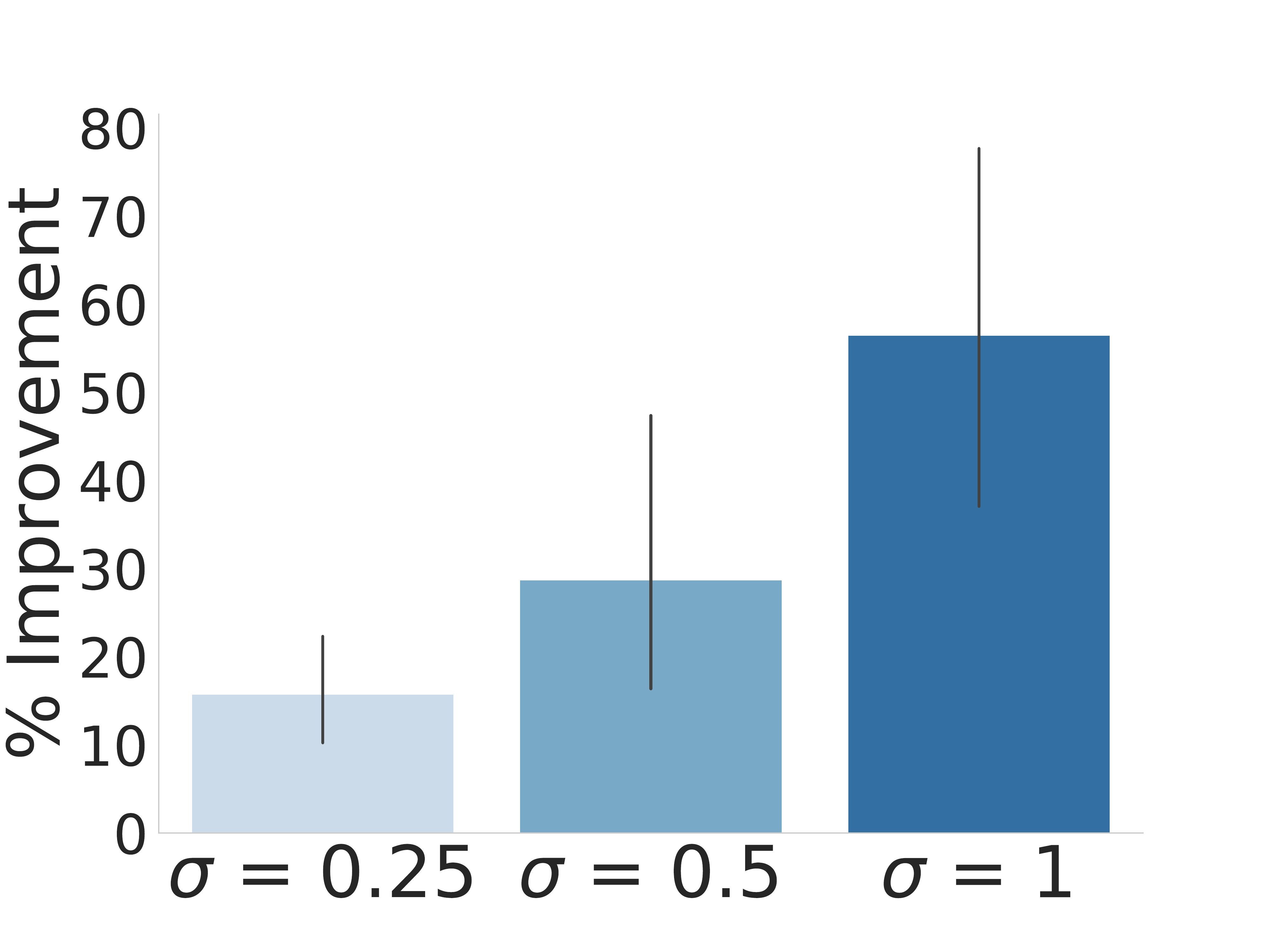}
  \caption{\small Improvements in robustness when the triangle leaf classifier is {\em renormalized}.}
  \label{fig:triangle_renormalize}
\end{figure}

For the remainder of this subsection, note that we measure the improvements in robustness for a small sample of triangular road sign inputs ($\sim$ 200). This is the reason for the increased improvement, as in comparison to those in \S~\ref{expts} where the gains are reported for over $\sim$ 3000 inputs. The results are presented in Figure~\ref{fig:triangle_retrain} and Figure~\ref{fig:triangle_renormalize}. Whilst providing comparable gains (on average), we reiterate the detriment caused by retraining approaches in Appendix~\ref{app:individual_circles} (refer Figure~\ref{fig:improve_retrain}).

\subsection{Retraining vs. Renormalization for Circular Road Signs}
\label{app:circles}

\begin{figure}[ht]
  \centering
  \includegraphics[width=0.5\linewidth]{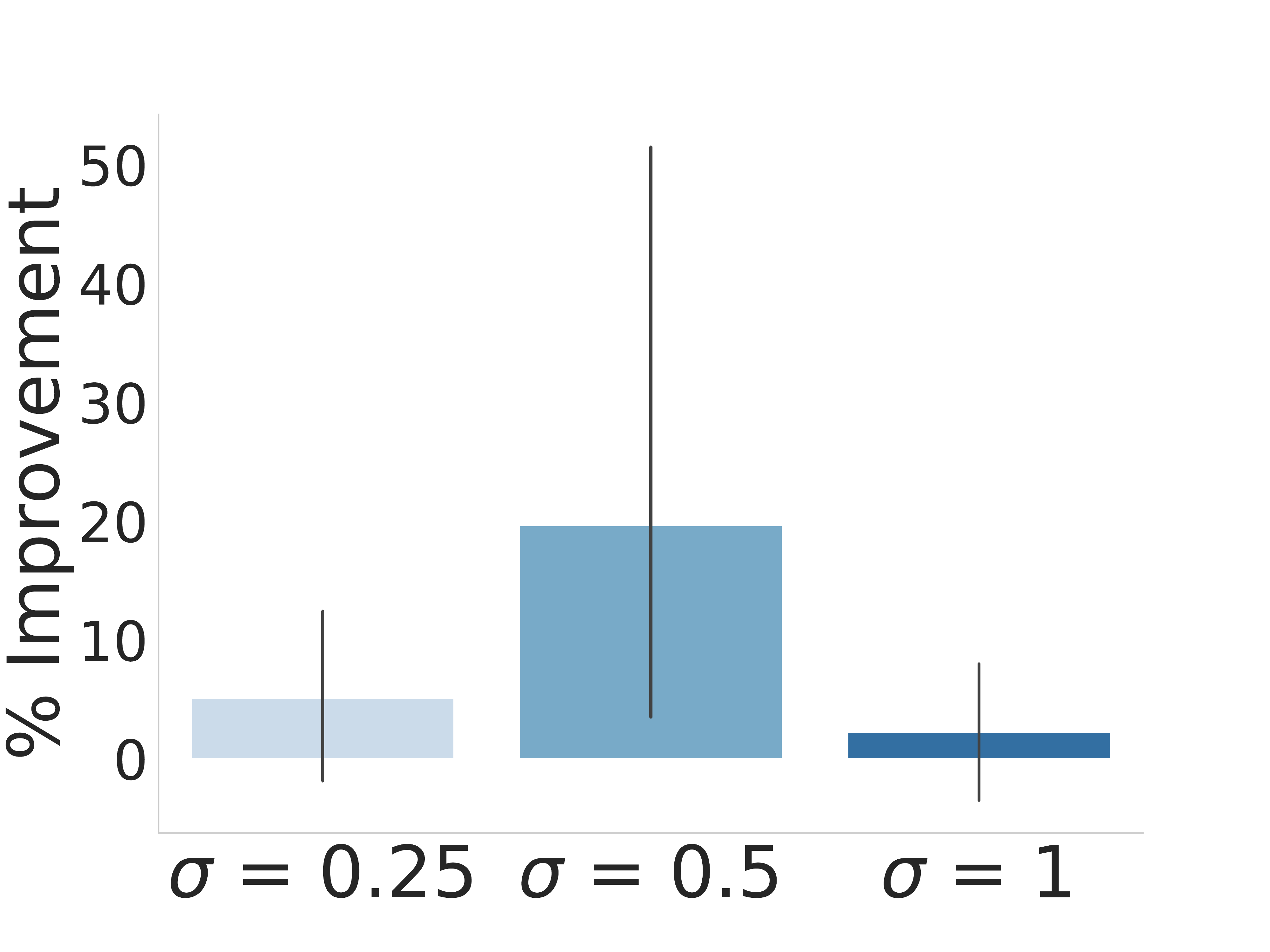}
  \caption{\small Improvements in robustness when the circle leaf classifier is {\em retrained}.}
  \label{fig:circle_retrain}
\end{figure}

\begin{figure}[ht]
  \centering
  \includegraphics[width=0.5\linewidth]{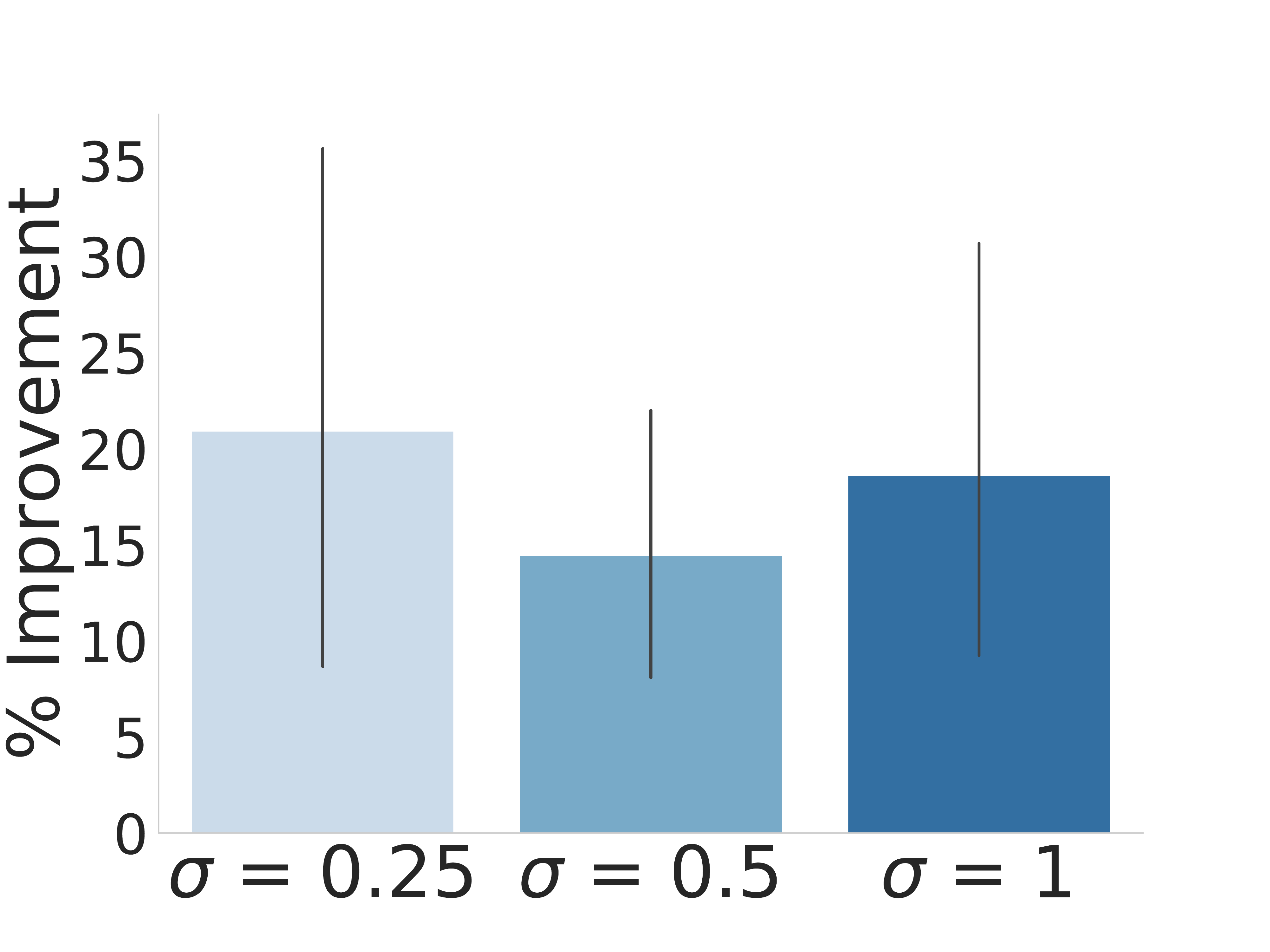}
  \caption{\small Improvements in robustness when the circle leaf classifier is {\em renormalized}.}
  \label{fig:circle_renormalize}
\end{figure}

For the remainder of this subsection, note that we measure the improvements in robustness for a small sample of circular road sign inputs ($\sim$ 200). This is the reason for the increased improvement, as in comparison to those in \S~\ref{expts} where the gains are reported for over $\sim$ 3000 inputs. The results are presented in Figure~\ref{fig:circle_retrain} and Figure~\ref{fig:circle_renormalize}. In comparison to the triangular road sign case, the improvements are relatively lower. This is because the runner-up candidates in the baseline (non-hierarchical classifier) are the same as the runner-up candidates for the hierarchical classifier. Thus, there is no significant widening in the margin. This further motivates the need to understand invariances associated with classification tasks, and obtain invariances that, by design, maximize this margin.

\subsection{Analysis of Individual Circular Road Signs}
\label{app:individual_circles}

\begin{figure}
  \centering
  \includegraphics[width=\linewidth]{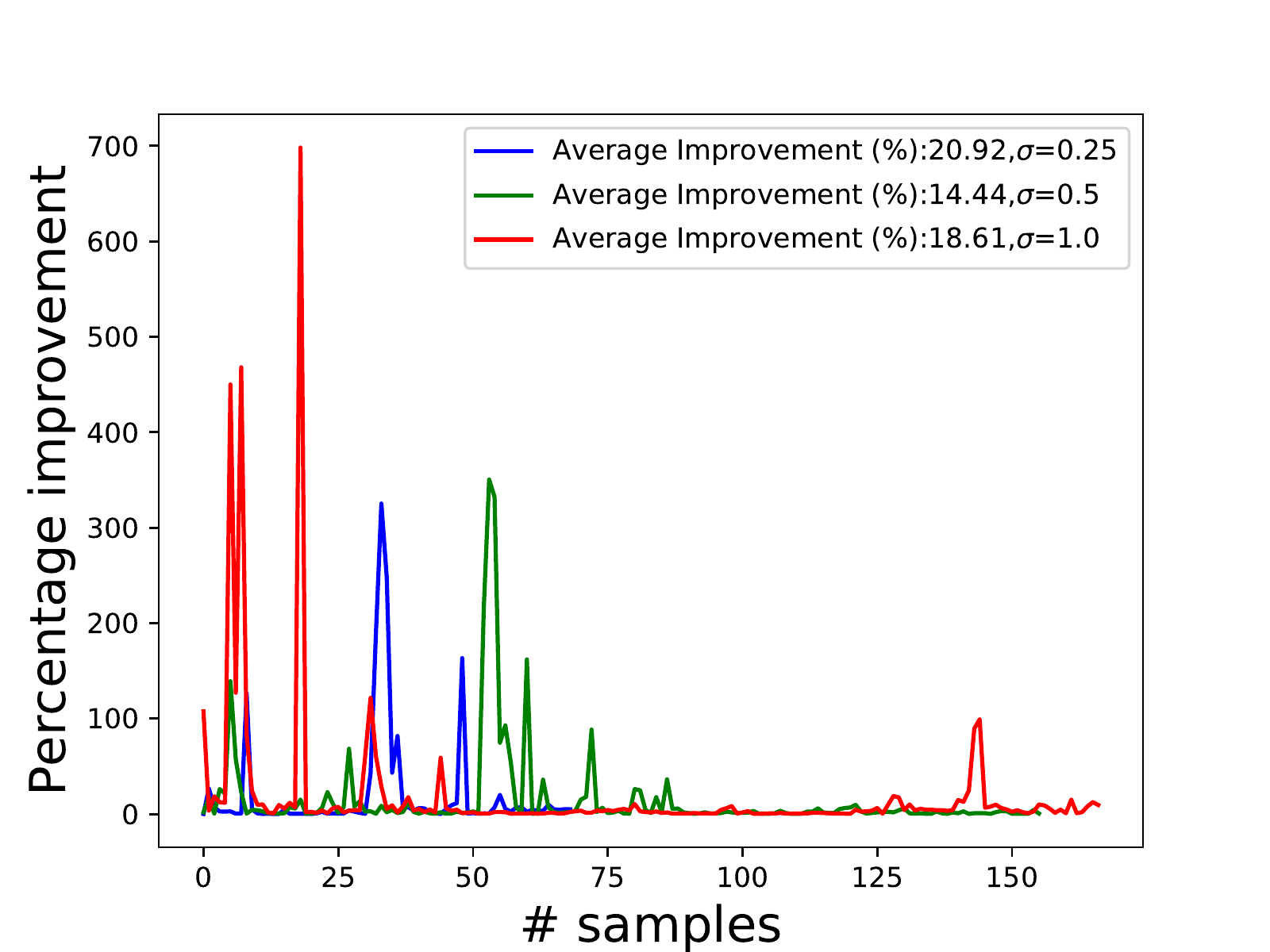}
  \caption{\small Individual robustness improvements when the leaf classifiers are {\em renormalized}.}
  \label{fig:improve_renormalize}
\end{figure}

\begin{figure}
  \centering
  \includegraphics[width=\linewidth]{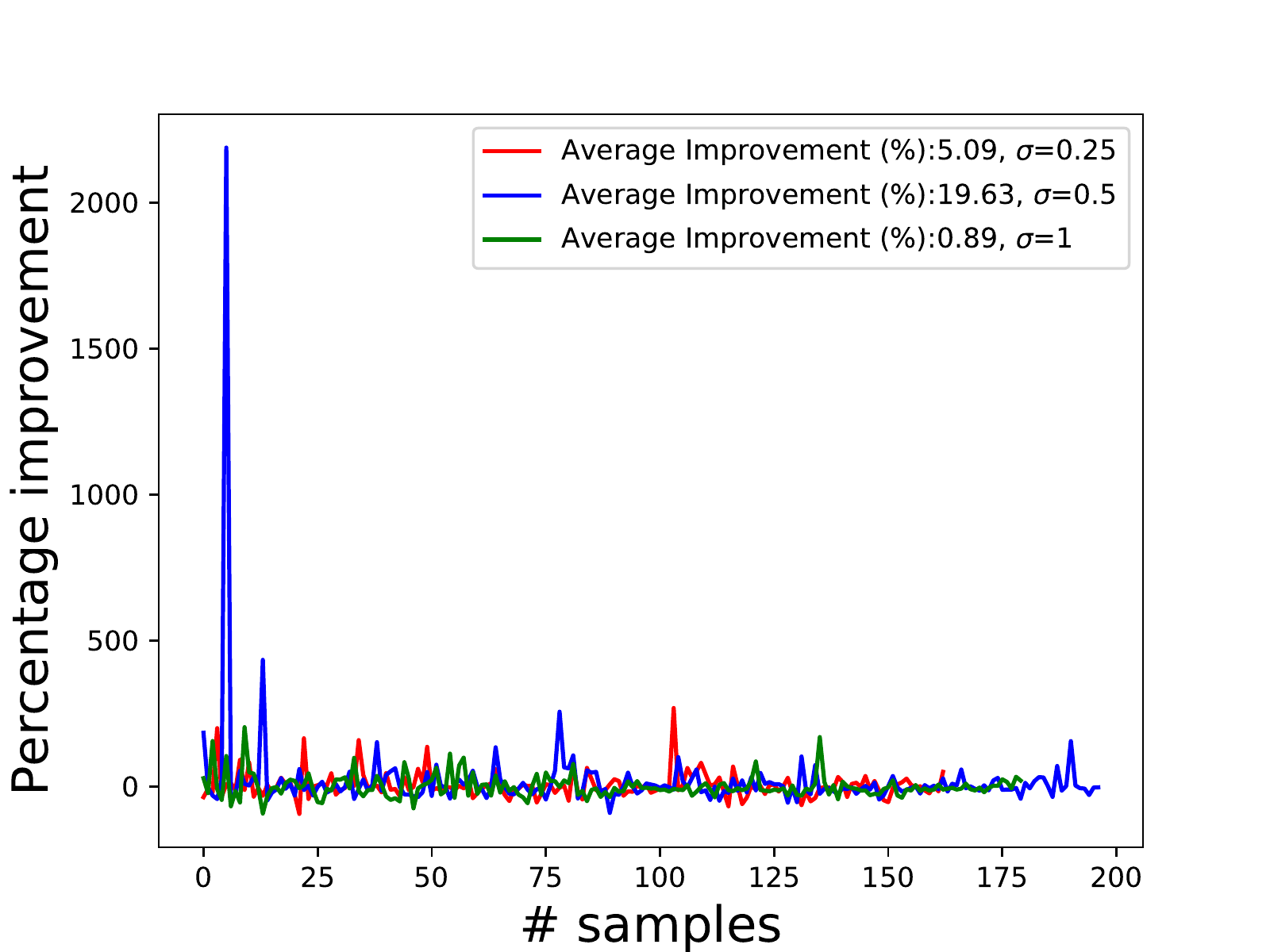}
  \caption{\small Individual robustness improvements when the leaf classifiers are {\em retrained}.}
  \label{fig:improve_retrain}
\end{figure}

We report fine grained measurements of the improvements discussed in Appendix~\ref{app:circles}. Here, we plot the percentage improvement (or degradation in the case of retraining) of robustness for each individual point. Observe that for some points, the increase in robustness is almost $7\times$ in the renormalization scenario (refer Figure~\ref{fig:improve_renormalize}), and about $20\times$ in the retraining scenario (refer Figure~\ref{fig:improve_retrain}). However, for most others, the increase in robustness is nominal. However, observe that in the retraining scenario, there are lots of points where the robustness decreases, which is why we utilize the renomralization approach for our experiments in \S~\ref{expts}.

%% file: ccs-2019/new_madry_stuff.tex
They work in a binary classification setting and have $d+1$ features.
The data distribution $\mathcal{D}$ is defined as follows: 
A sample $(x,y)$ is generated as follows:
\begin{itemize}
\item First a label $y$ is sampled uniformly from $\{-1,+1 \}$. 
\item Feature $x_1$ takes on value $+y$ with probability $p$ and
$-y$ with probability 1-p.
\item Features $x_2,\cdots,x_{d+1}$ are sampled from the distribution
$\N(\eta y,1)$
\end{itemize}
$\N(\mu, \sigma^2)$ is the normal distribution with mean $\mu$ and standard deviation $\sigma$. 
This setting defines a binary classification task with two types of features: one robust feature that predicts the correct label with a probability $p$ and a set of features that ``moderately'' predict the output label sampled from $\N(\eta y,1)$. A classifier that averages the features $x_2,\ldots,x_{d+1}$ has near perfect natural accuracy but fails under an $\ell_\infty$-bounded adversary that can perturb each feature by $2\eta y$. Tsipras et al. formalize this observation by the following theorem (reproduced from their recent paper~\cite{tsipras2018there} 
with a simple change of notation to avoid clash of notation for $\delta$):

\begin{theorem}\label{thm:lunch}
Any classifier that attains at least $1-\gamma$ natural accuracy on $\mathcal{D}$ has adversarial accuracy at most $\frac{p}{1-p}\gamma$ against an $\ell_\infty$-bounded adversary with $\|\delta\|_{\infty} \geq 2\eta y$.
\end{theorem}

Theorem~\ref{thm:lunch} shows that, in this setting, a trade-off between robustness and accuracy exists. For example, a classifier with natural accuracy of $99\%$ has an adversarial accuracy of $19\%$ when $p=0.95$. The robust feature provides a baseline natural and adversarial accuracy equal to $p$. To improve the natural accuracy beyond $p$, the classifier relies on the other features. An $\ell_\infty$-bounded adversary that can perturb each feature by $2\eta y$ effectively flips the distribution of $x_2,\ldots,x_{d+1}$ from $\N(\eta y,1)$ to $\N(-\eta y,1)$. The result of a model trained on the original distribution is unreliable under this flipped distribution, which degrades the adversarial accuracy.


 
Imposing an invariant on the same $\ell_\infty$-bounded adversary prevents it from completely flipping the distributions of $x_2,\ldots,x_{d+1}$ from $\N(\eta y,1)$ to $\N(-\eta y,1)$. The invariant restricts the attacker's actions within the $\ell_\infty$ ball; a more restrictive invariant improves the adversarial robustness. Consider an invariant where the adversary is not allowed to perturb the first $k$ features. 

In such a case, one can construct a meta-feature $x'_1$ defined as the sign of the average of $x_2,\ldots,x_{k+1}$ such that: $x'_1 = \sign\left(\frac{1}{k}\sum_{i=2}^{k+1}{x_i}\right)$, where $\frac{1}{k}\sum_{i=2}^{k+1}{x_i} \sim \N(\eta y,\frac{1}{k})$. The meta-feature predicts $y=1$ when $x'_1\geq 0$ and predicts $y=-1$ when $x'_1 < 0$.

This meta-feature is predictive of $y$ as follows: 
\begin{equation}
\begin{split}
  p' = P(x'_1=y)= \\
  \frac{1}{2}P\left(\N\left(\eta y,\frac{1}{k}\right)\geq 0 \,\middle|\, y=1\right) + \\
  \frac{1}{2}P\left(\N\left(\eta y,\frac{1}{k}\right)<0 \,\middle|\, y=-1\right)\\
    = P\left(\N\left(\eta,\frac{1}{k}\right)\geq 0 \right)
    = 1 - \Phi^{-1}(-\sqrt{k}\eta).
    \end{split}
\end{equation}
Recall that $\Phi^{-1}$ is the inverse of the CDF for the normal distribution. 
If $k$ is large enough such that $k \geq 9/\eta^2$, then $p'>0.99$. Recall that as long as $k < d$, this adversary is still the same $\ell_\infty$-bounded adversary with which we started the discussion. Then, $x'_1$ is a robust feature that is highly predictive of the output. This meta-feature was constructed via imposing a constraint on the adversary, and is not similar to $x_1$ which is provided as part of the problem setup.

Given the result of theorem~\ref{thm:lunch}, the robustness-accuracy trade-off can reworded as: \textit{Any classifier that attains at least $1-\gamma$ natural accuracy on $\mathcal{D}$ has adversarial accuracy at most $\frac{p'}{1-p'}\gamma$ against an $\ell_\infty$-bounded adversary with $\|\delta\|_{\infty} \geq 2\eta y$ subject to the invariant that the first $k$ elements of $\delta$ are zero.} If the value of $k$ is large enough, the value of $p'$ can be much larger than that of $p$ so that both natural accuracy and adversarial accuracy can be close to $99\%$.

\begin{figure}[t]
  \centering
  \includegraphics[width=\linewidth]{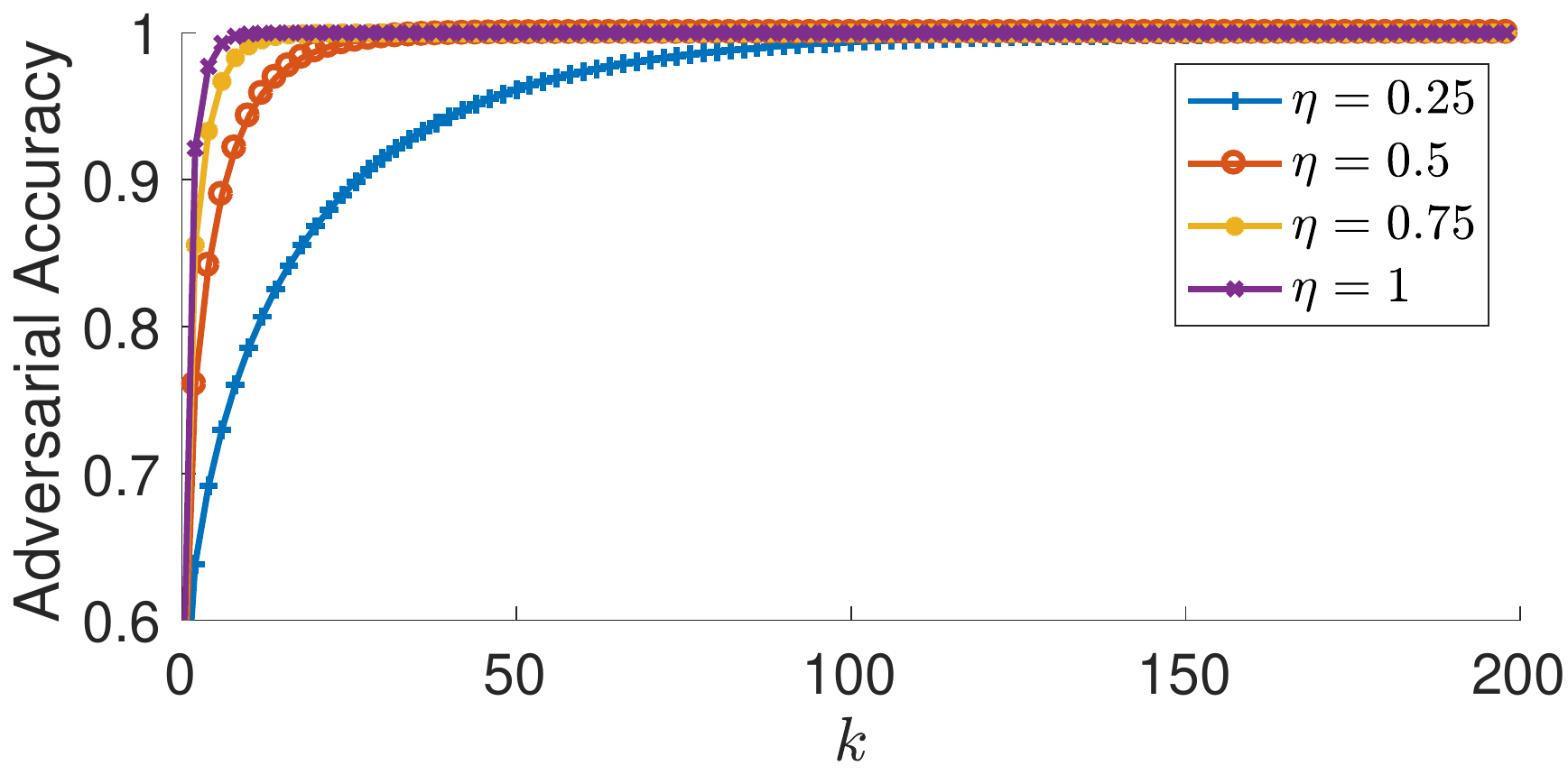}
  \caption{The improvement in adversarial accuracy as the invariant imposes a tighter constraint in the setting of Tsipras et al.~\cite{tsipras2018there}.}
  \label{fig:madry_examples}
\end{figure}

Figure~\ref{fig:madry_examples} shows the adversarial accuracy for a toy example with $d=200$ and for different values of $\eta$ less than or equal to 1. For all the values of $\eta$, the natural accuracy reaches nearly 100\%. The value of $k$ controls the invariant. It is evident from the figure, that for small values of $\eta$, even when $k$ is much smaller than $d$, the adversarial accuracy exceeds 90\%. The main takeaway is that imposing an invariant even on a relatively small number of features can improve the defender's adversarial robustness. 